\documentclass[letterpaper, conference]{IEEEtran}
\usepackage[letterpaper, top=57pt, bottom=54pt, left=50pt, right=50pt]{geometry}
\usepackage{cite}
\usepackage{amsmath,amssymb,amsfonts}
\usepackage{algorithmic}
\usepackage{graphicx}
\usepackage{textcomp}
\usepackage{xcolor}
\usepackage{graphicx}
\usepackage{subcaption}
\usepackage{overpic}
\usepackage{makecell}
\usepackage{booktabs}
\usepackage{adjustbox}
\usepackage{multirow}
\def\BibTeX{{\rm B\kern-.05em{\sc i\kern-.025em b}\kern-.08em
    T\kern-.1667em\lower.7ex\hbox{E}\kern-.125emX}}

\begin{document}

\title{TacClip: a clip-on sensor measures dynamic contact forces without covering the fingerpads}
%MRC: It's not really about pressure. How about this^^. It seems like this captures the key selling points.
%\hao{I'm suggesting a more precise tile: TacClip: Clip-on FBG Sensing for Integrated Quasi-static Pressure and Dynamic Vibration Measurement}


% \author{Author names removed for double-anonymous review}

\author{
    \IEEEauthorblockN{
        Yuqian Ye\textsuperscript{1}, 
        Hao Li\textsuperscript{1}, 
        Jingxi Xu\textsuperscript{1, 2}, 
        Haojun Feng\textsuperscript{1}, 
        Seongheon Hong\textsuperscript{1}, 
        Mark R. Cutkosky\textsuperscript{1} 
    }
    \IEEEauthorblockA{
        \textsuperscript{1}Department of Mechanical Engineering, Stanford University, Stanford, CA, USA \\
        \textsuperscript{2}Ant Group, Sunnyvale, CA, USA \\
        Email: \{yuqianye, li2053, hfeng2, wbfprp,  cutkosky\}@stanford.edu, j.x@antgroup.com
    }
}

     \maketitle

\begin{abstract}
TacClip is a minimally encumbering wearable device for recording fingertip deformation caused by contact forces and vibrations. It can be combined with vision- or glove-based hand tracking systems that leave the fingertips uncovered and provides a measure of dynamic contact interactions, while leaving the finger pads exposed  so that the user retains natural sensitivity to texture, friction, temperature, and fine surface features. The signal is produced by a Fiber Bragg Grating (FBG) embedded on a small plastic clip mounted over the fingernail. Optionally, for use with vision-based tracking, additional FBGs on polyimide strips can complement camera-based pose estimation. In finger pressing tests, TacClip estimates the force magnitude with typical errors below $0.5~\mathrm{N}$ over a $0$--$8~\mathrm{N}$ range. In tests of cloth handling and tape edge finding, we show that it captures the vibrations and dynamic events generated during exploratory sliding.  With no electronics, TacClip can also be used submerged in water, while preserving bare finger contact. 
%MRC Now at 1072 words.
%\hao{the abstract seems over the limit of RAL requirements (1200 words)}
\end{abstract}

\begin{IEEEkeywords}
tactile sensing, wearable sensing, Fiber Bragg Grating, texture recognition
\end{IEEEkeywords}

% \hao{file size is 9MB, limit = 5MB} \ivy{resize all figures, the file should be smaller than 5MB}
%MRC We can check graphics or compress PDF I think

\section{Introduction}
Robot manipulation policies increasingly rely on large collections of human demonstrations. Most demonstration interfaces emphasize visual observations and hand or tool trajectories, but many manipulation tasks are fundamentally contact-rich: success depends on when contact is made, how force is modulated, and what tactile events occur during sliding, pressing, or grasping. Recent surveys and data-collection systems reflect this trend toward force-aware and touch-aware robot learning \cite{xie2025forceful,suomalainen2022survey,chi2024universal,wang2024dexcap,liu2025forcemimic,yu2024mimictouch}. However, capturing the contact information of a human demonstrator remains difficult, because the sensing system has to retain the tactile cues that guide the human hand.

\begin{figure}[t]
    \begin{overpic}[trim=800pt 0pt 500pt 1000pt, clip=true, width=0.98\columnwidth]{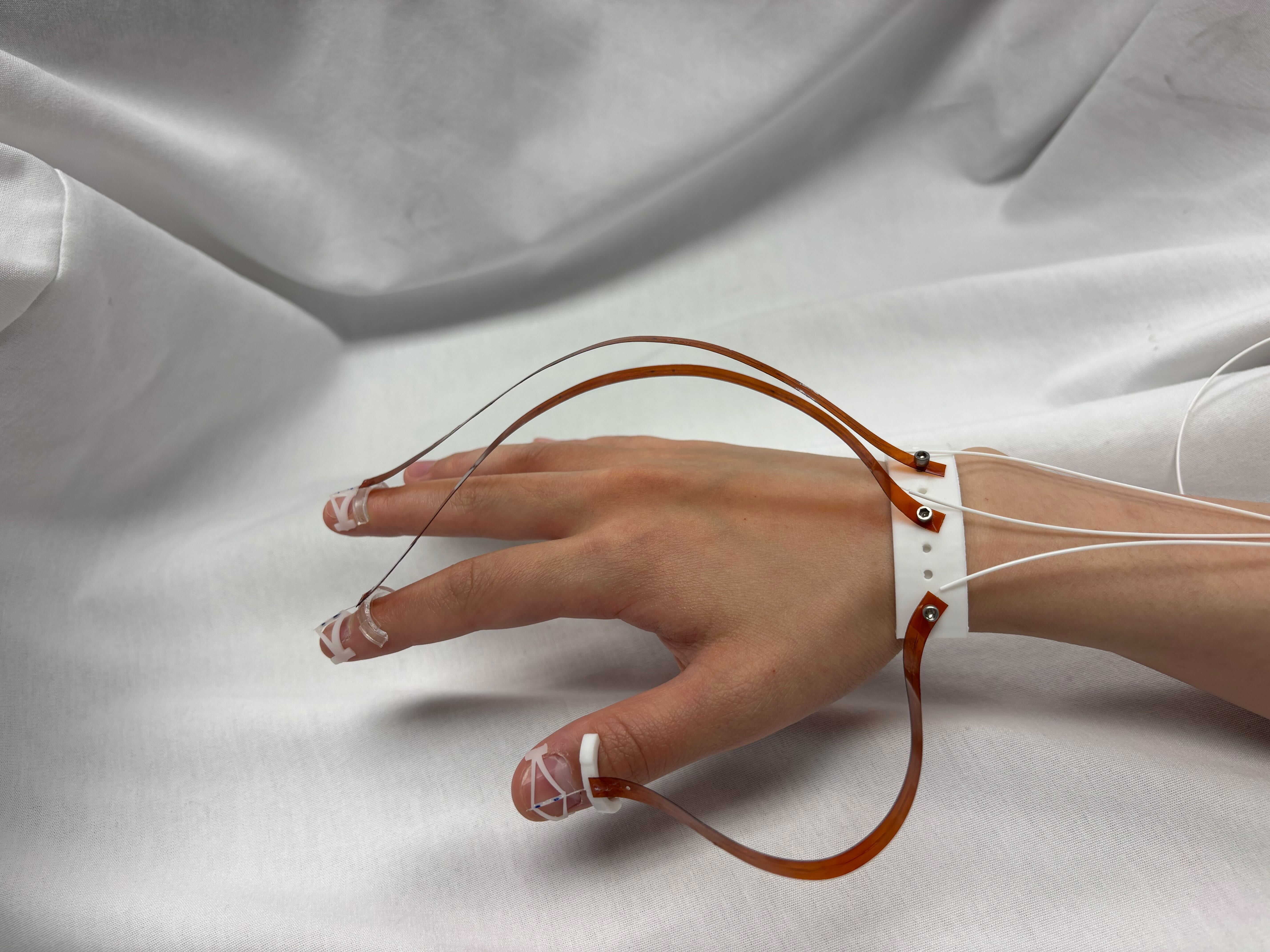}
        \put(1,1){\includegraphics[width=0.25\columnwidth]{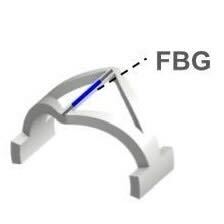}}
    \end{overpic}
    \caption{TacClip mounted on a fingertip
with the optical fiber routed over a dorsal polyimide strip, attached to
a wristband. 
%\jx{It could be nice to have a zoomed-in shot of just the thumb area with TacClip, overlayed on top at the corner of this image. \mc{Agreed. Also see question in source}}
}
%mrc Ivy: Are the screws that fasten the kapton strips to the wristband tight, or do they permit rotation? (this obviously affects how the FBGs on the strips correspond to finger motion)
%Ivy Answer: they permit rotation. the screws sre designed for no-torsion on the tape, allow the tape to follow the finger's in-plane rotation as much as possible.

    \label{fig:tacclip_design}
\end{figure}

Several approaches can measure or estimate fingertip contact. Pressure sensors can be attached to the fingertip or embedded in a glove; exoskeletal devices can estimate contact from structure-mounted force sensors; and vision-based approaches can infer forces from observed deformations. For recent surveys, see \cite{belcamino2024systematic, shao2026manufacturing,li2024systematic,ko2023vision}---and new developments are announced regularly.
%\jx{It would be nice to place references right after these parallel clauses.} 
%\mc{yes - if we can find a good recent survey that makes it easy.}
These methods are valuable, but many of them place material between the fingerpad and the environment, restrict dexterity, or require bulky fixtures. Such encumbrance is a serious limitation for dexterous activities in which the human demonstrator relies on bare-finger cues, for example, feeling a fabric ridge, locating a transparent edge, or palpating a compliant surface.

TacClip belongs to a complementary class of wearable sensors that seek to measure contact indirectly while leaving the fingerpad exposed. The device clips around the fingertip rather than covering the pad. Contact loads deform the fingertip and produce small elastic deformations in the clip; an embedded FBG converts those deformations into wavelength shifts. The same optical fiber can carry additional FBGs along a thin dorsal strip, enabling curvature measurements that can augment visual hand-pose tracking without requiring a full glove.

The contributions of this paper are threefold. First, we introduce a clip-on FBG fingertip sensor that measures contact-induced fingertip deformation while preserving direct bare-finger contact. Second, we characterize the sensor for both 
%\jx{Can you explain to me why "quasi-static"?} 
%mrc strictly speaking, static is stationary. quasi-static means moving, but slowly enough that no inertial or damping terms need be considered. I'm OK with either term in this case. If data were taken while fingertip load was changing, it's quasistatic. If data were taken for static loads, one at a time, it's static.
quasi-static contact-force magnitude estimation and dynamic vibration sensing, including the frequency response of the fingertip--TacClip system under swept-frequency loading. Third, we demonstrate applications that are difficult for vision or conventional covered-fingertip gloves alone: directional cloth-texture discrimination, transparent tape-edge detection, and underwater force-response measurement. We also discuss how the optional FBG curvature strip can be integrated with visual hand tracking for occlusion-prone manipulation and palpation tasks.

\section{Related Work}
\subsection{Tactile and Force Information for Robot Learning}
Tactile and force feedback have long been recognized as essential for stable manipulation, slip detection, texture perception, and contact-rich control. Recent robot-learning systems have renewed interest in collecting richer human demonstrations because purely visual or kinematic demonstrations can omit forces that are central to task success \cite{xie2025forceful,suomalainen2022survey}. Data-collection interfaces such as UMI~\cite{chi2024universal} and DexCap~\cite{wang2024dexcap} enable scalable human demonstration capture for manipulation, mainly through portable visual, pose, and motion-capture pipelines. More force- and touch-centered systems, including MimicTouch~\cite{yu2024mimictouch}, ForceMimic~\cite{liu2025forcemimic}, and UME~\cite{liang2026universal} show the value of tactile or force information for transferring human strategies to robot policies. TacClip addresses a sensing gap in this space: it is intended to record tactile events during natural bare-finger demonstrations without requiring a handheld gripper, a thick glove, or a sensor layer under the fingerpad.

\subsection{Wearable Fingertip and Glove Sensors}
Wearable tactile gloves can record rich human contact patterns. The scalable tactile glove of Sundaram \emph{et al.} demonstrated dense grasp signatures across the hand, while digitally embroidered smart gloves have shown scalable textile-based tactile sensing and haptic actuation for recording and transferring tactile interactions \cite{sundaram2019tactileglove,luo2024smartgloves}. These systems are powerful for large-area hand sensing, but glove layers and embedded sensing elements can attenuate bare-skin perception or alter friction at the contact interface. Fingertip-mounted devices such as ThimbleSense~\cite{battaglia2016thimblesense} provide detailed contact and force information for grasp analysis, but they still instrument the contact surface itself. TacClip instead measures deformation around the fingertip while leaving the fingerpad uncovered.

Some prior systems also leave the fingerpad free by sensing through the nail. Mascaro and Asada used photoplethysmographic fingernail sensors to estimate fingertip forces without obstructing haptic contact, and later extended the approach to estimate finger posture and three-axis fingertip force \cite{mascaro2001photoplethysmograph,mascaro2004fingernail}. More recent nail-mounted strain sensors measure fingernail deformation for biomechanical monitoring and human-computer interaction \cite{sakuma2018naildeformation,hsiu2016nailplus}. These nail-based systems are closely related in motivation because they exploit distal-finger deformation rather than a force sensor under the pad. TacClip takes a different mechanical route: it avoids nail mounting and instead uses a removable clip on the fingertip, enabling fast installation and straightforward integration with optical-fiber multiplexing.

\subsection{FBG-Based Tactile and Hand-Posture Sensing}
FBG sensors are attractive for wearable and robotic tactile sensing because they are lightweight, immune to electromagnetic interference, highly sensitive to strain, and can be multiplexed along a single fiber \cite{hill1997fbg,kersey1997fibergrating}. These properties have motivated FBG sensing gloves for monitoring finger posture and joint flexion \cite{dasilva2011fbgglove,jha2021fbgglove}. FBGs have also been embedded in robot fingers and tactile structures for contact detection and force control \cite{park2007fbgfinger,park2008fingertipforce,heo2006tactilearray}. 
%MRC Sorry, but I think this is too off-topic:
% More recently, FBG-based whisker sensors have been used for underwater tactile contact tracking, illustrating the suitability of optical strain sensing for aquatic environments \cite{li2024whisker}. 
TacCap~\cite{xing2025taccap}, a recent wearable FBG tactile sensor, further demonstrates the promise of FBG sensing for human-to-robot skill transfer. TacClip shares the same broad motivation of FBG-based wearable tactile data collection, but its clip-on geometry is designed specifically to keep the fingerpad exposed and to capture both static deformation and dynamic vibration from the user's own fingertip.

\subsection{Dynamic Touch Sensing}
%mrc I actually don't think we need this paragraph. These are robotic sensors, not for human wearing.
% Vision-based optical tactile sensors such as GelSight, TacTip, and DIGIT provide high-resolution contact geometry and force-related information for robot hands \cite{yuan2017gelsight,wardcherrier2018tactip,lambeta2020digit} 
% %\jx{my personal preference is to always put citation right after the name, such as GelSight [1], DIGIT [2], etc.}. 
% These sensors typically require a deformable optical skin to contact the object, which is appropriate for robot end-effectors but not for recording human demonstrations in which the human should feel the object directly. TacClip uses optical sensing in a different sense: the optical element is a strain-sensitive fiber that is outside the primary contact interface.

Dynamic tactile cues are particularly important for texture exploration and slip perception. Human haptic exploration uses characteristic exploratory procedures, including lateral motion for texture assessment \cite{lederman1987hand}. Robotic (e.g. 
\cite{howe1989sensing, chen2018tactile,feng2019slip,adachi2026ai}) and biological (e.g. \cite{johansson2009coding}) studies have likewise shown that skin acceleration and vibration contain information about slip and surface texture.

TacClip is designed to preserve the user's natural exploratory motion while recording the resulting contact-induced vibration at high sampling rate. This capability is relevant to fabric manipulation, where tactile feedback can disambiguate fine structures such as layers, ridges, or weave direction that are difficult to infer from vision alone \cite{tirumala2022cloth}.

% The device described in this paper is new, but draws upon prior work in the use of optical fibers with FBGs for sensing in manipulation and on previous efforts to develop unenumbering contact sensors.

% Cite Korean and other FBG-based glove, exoskeltal things. 

% Also TacCap. Check TacCap paper for more citations

% An early example is the photopethysmograph by Mascaro and Asada [cite] and a more recent example is the fingernail 

% A much newer solution is described in [fingernail strain gages] – 

\section{Mechanical Design of TacClip}
\label{sec:design}

\begin{figure}[htb]
    \centering

    \begin{subfigure}{\columnwidth}
        \begin{overpic}[trim=70pt 240pt 0pt 90pt, clip=true, width=0.98\columnwidth]{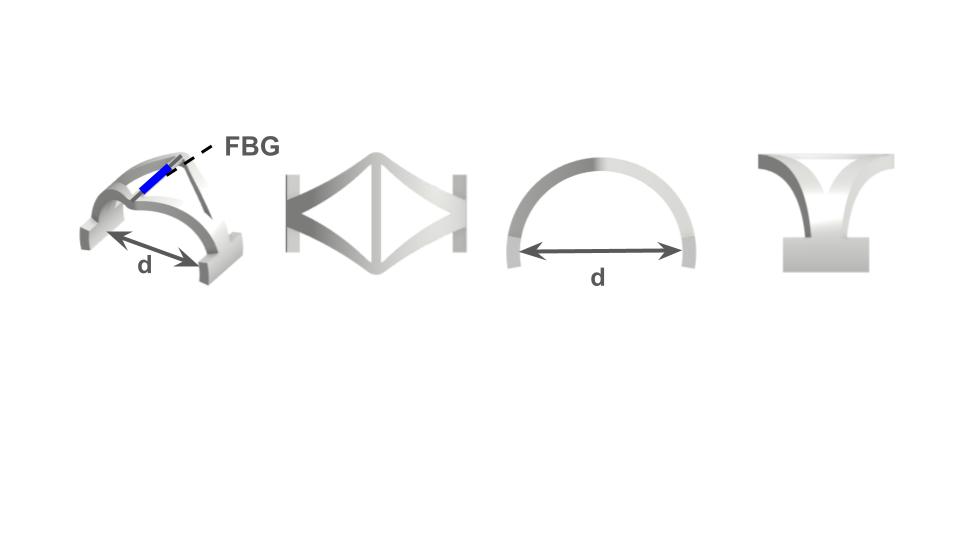}
            \put(2,18){{\textbf{(a)}}}
        \end{overpic}
    \end{subfigure}

    \begin{subfigure}{0.98\columnwidth}
        \begin{overpic}[trim=10pt 10pt 0pt 70pt, clip=true, width=0.98\columnwidth]{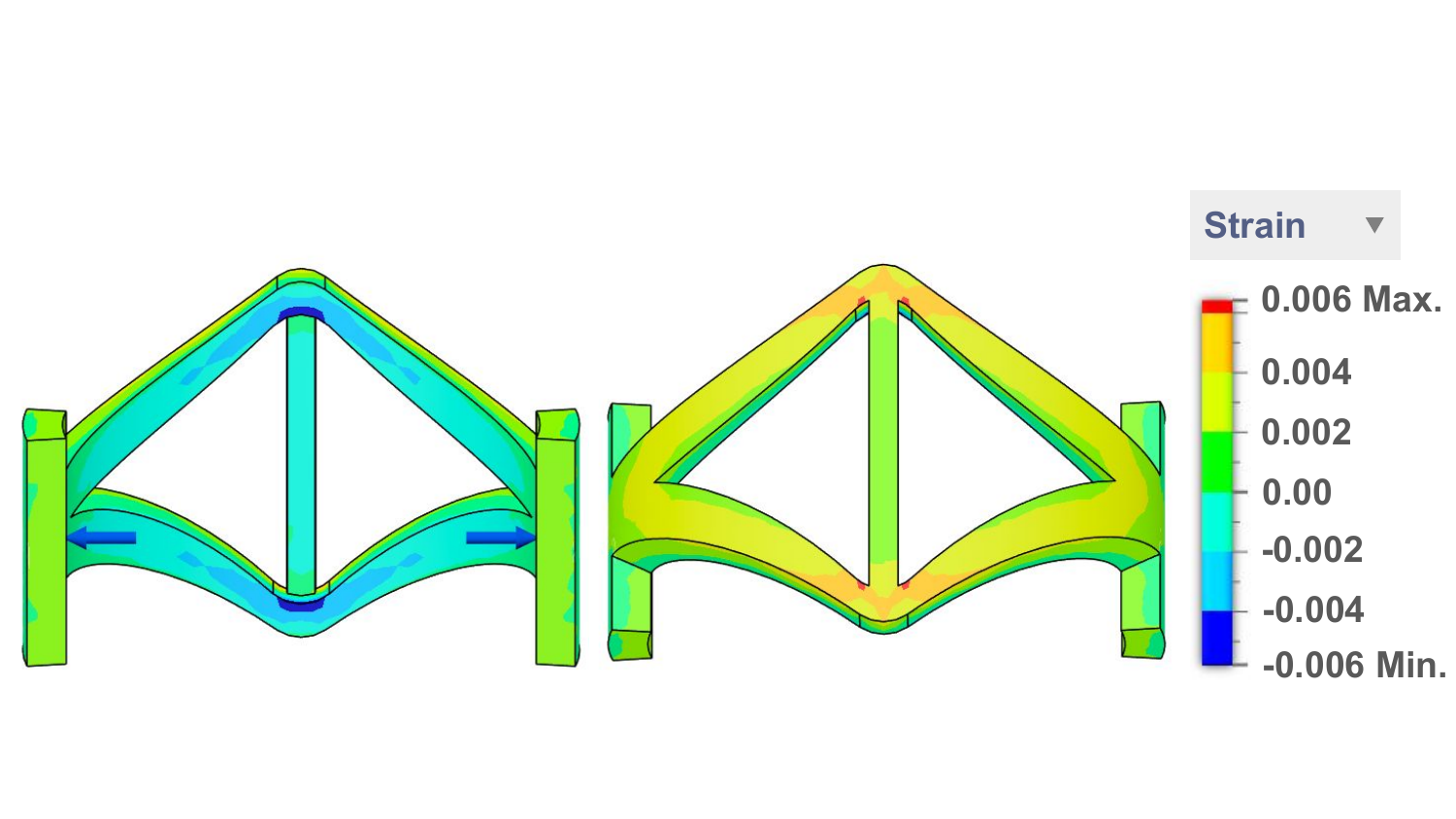}
            \put(2,35){{\textbf{(b)}}}
        \end{overpic}
    \end{subfigure}

    \caption{TacClip CAD design and FEA result. (a) CAD model of the clip. (b) Simulated strain on the outer and inner surfaces of the central strut when finger lateral expansion applies normal forces to the inner contact surfaces (blue arrows). 
   % \mc{Let's make the CAD ad FEA bigger (see note)}
    %MC I think the CAD and FEA analysis to show the predicted stress as a function of loading (and how it matches what we actually measure) is important enough to make this bigger. This shows that we (i) used FEA to improve (not formally optimize) the design and (ii) we get the stresses in the FBG that we expect. 
   % \mc{Maybe make it another figure. We have room.}
    %\jx{Colorbar text is still too small to see. My rule of thumb, always for paper figures, is that the smallest text in the whole paper should not be smaller than footnote size.}
%    \mc{Ideally show double arrow to clarify the size: $d$}
    }
       
    \label{fig:tacclip_CAD_FEA}
    
\end{figure}

% The fingertip unit of TacClip is 3D printed from Rigid 4000 resin and is sized to fit snugly around the distal fingertip. Different clip sizes are used for small, medium, and large fingers. Unlike adhesive strain gauges or pressure pads, the clip does not require glue or tape on the skin. A single-mode optical fiber containing an FBG is bonded to the central beam of the clip. When the fingertip is loaded, the soft tissue bulges laterally and bends the clip, producing strain in the fiber and shifting the reflected Bragg wavelength.
% Replace XX, YY, ZZ, AA, BB, CC, and TT with measured values.
The fingertip unit of TacClip is 3D printed from Rigid 4000
resin and is donned by sliding it over the distal fingertip.
It requires no adhesive or tape and leaves the
fingerpad uncovered. We prepared three nominal inner diameters
of $d$ = (12, 13, 14.4)~mm (small, medium, and large), covering
the tested distal fingertip-width range of 13.5--17~mm;
intermediate sizes can be generated by scaling the CAD
geometry. A snug, repeatable fit is important because clip
preload affects the wavelength baseline and force calibration.
For users between the nominal sizes, thin removable high-friction
silicone inserts, such as eyeglass nose-pad material, can be
placed on the inner side surfaces to reduce slip without
covering the fingerpad. The size and insert configuration
are selected before calibration; visible clip motion requires
a baseline reset or recalibration. A single-mode optical fiber containing an FBG is bonded to
the central beam of the clip (Fig.~\ref{fig:tacclip_CAD_FEA}a).
At prototype scale, the
printed clip costs approximately \$2, and the complete
clip-and-FBG sensor head costs approximately \$20 per finger,
excluding assembly labor and the external optical interrogator.

When the fingertip is loaded,
the soft tissue bulges laterally and bends the clip 
as shown in Fig.~\ref{fig:tacclip_CAD_FEA}b, producing
strain in the fiber and shifting the reflected Bragg wavelength.
The operating principle follows the standard FBG relationship
\begin{equation}
    \lambda_B = 2 n_{\mathrm{eff}} \Lambda,
\end{equation}
where $\lambda_B$ is the reflected Bragg wavelength, $n_{\mathrm{eff}}$ is the effective refractive index of the fiber core, and $\Lambda$ is the grating period. Strain and temperature perturb $n_{\mathrm{eff}}$ and $\Lambda$, producing a measurable wavelength shift \cite{hill1997fbg,kersey1997fibergrating}. In TacClip, the dominant short-term signal during contact is strain induced by fingertip deformation; temperature drift is slower and can be compensated or interpreted separately, as discussed in the underwater experiment.

Fig.~\ref{fig:tacclip_design} shows the complete prototype, including the optional dorsal strip routed to a wristband. The CAD model in Fig.~\ref{fig:tacclip_CAD_FEA}(a) shows a detailed CAD design of the sensor, including the central strut that carries the FBG. Finite-element results in Fig.~\ref{fig:tacclip_CAD_FEA}(b) show that fingertip loading produces opposite strain signs on the outer and inner surfaces of the strut, consistent with bending. 
The clip geometry makes it sensitive to loads that produce lateral expansion of the fingertip.

If desired, the same optical fiber can extend beyond the clip and be bonded to a thin polyimide strip along the dorsum of the finger and hand, as shown in Fig.~\ref{fig:tacclip_design}. In the current prototype, the strip is $6~\mathrm{mil}$ ($0.15~\mathrm{mm}$) Kapton, and the fiber is bonded with Loctite 401 adhesive. Additional FBGs along this strip measure local curvature. These curvature signals do not replace full hand-pose tracking, but they can provide useful flexion information during occlusions, especially when fused with vision-based skeleton estimates. This capability is used in Section \ref{sec:augmenting}.

\section{Characterization of TacClip}
\label{sec:characterization}

\begin{figure}[t]
    \centering
    \begin{minipage}[b]{0.35\columnwidth}
        \centering
        \begin{overpic}[width=\linewidth, trim = 0 0 300pt 0, clip=true]{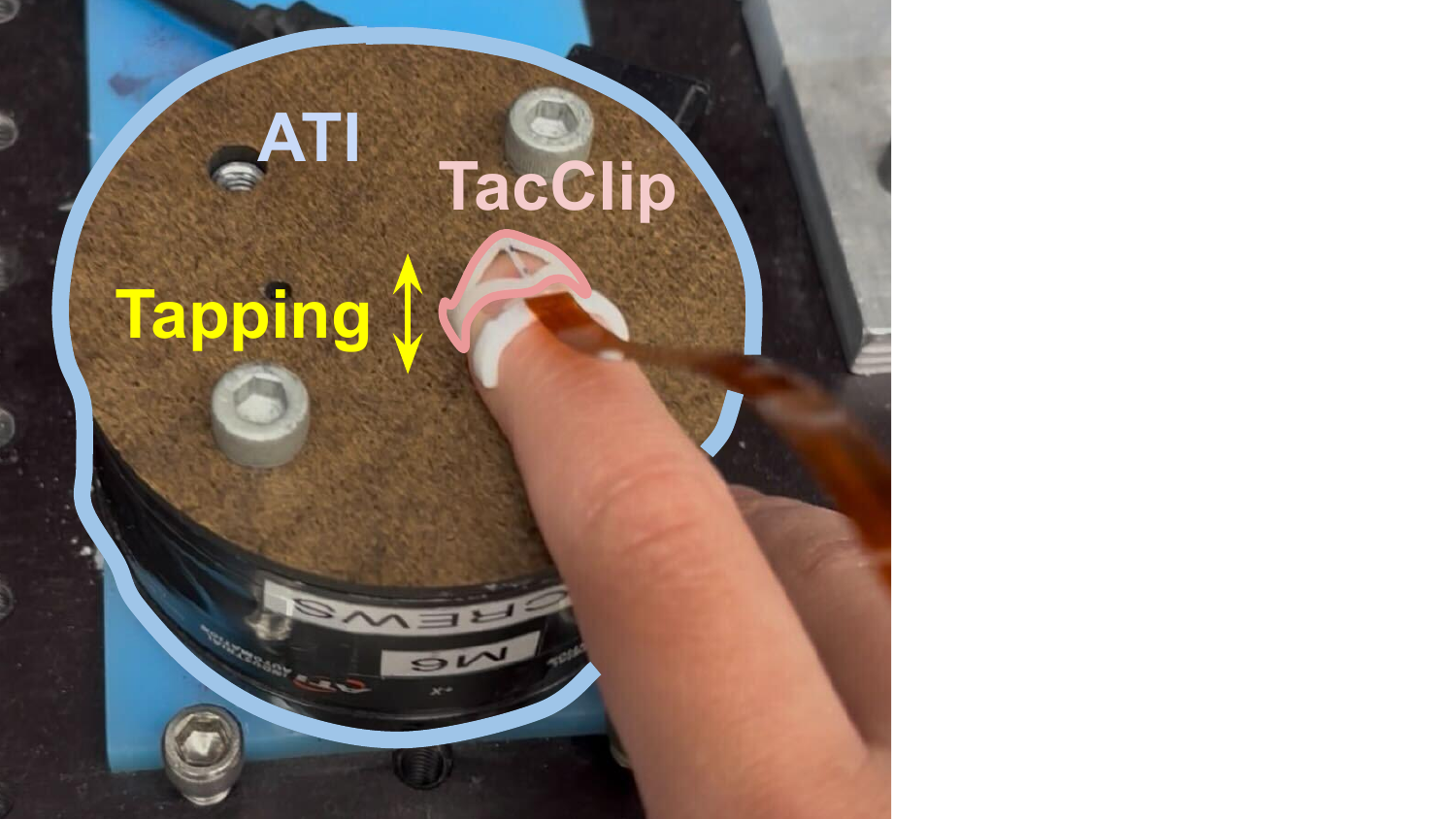}
            \put(5,5){\textcolor{white}{\textbf{(a)}}}
        \end{overpic}
    \end{minipage}
    \hfill
    \begin{minipage}[b]{0.63\columnwidth}
        \centering
        \begin{overpic}[width=\linewidth]{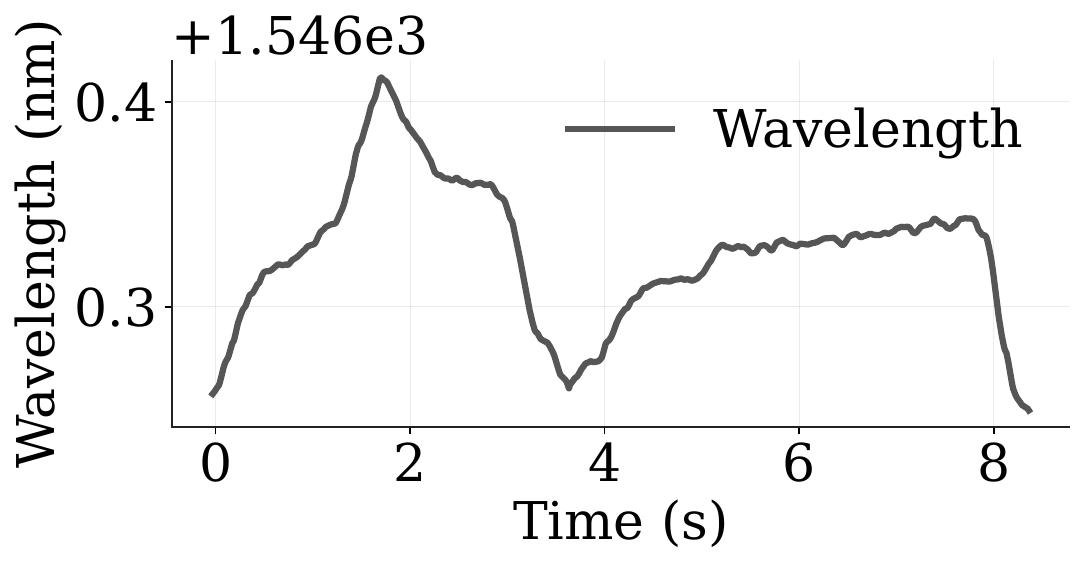}
            \put(5,5){\colorbox{white}{\textbf{(b)}}}
        \end{overpic}
    \end{minipage}
    \begin{minipage}[b]{0.98\columnwidth}
        \centering
        \begin{overpic}[width=\linewidth]{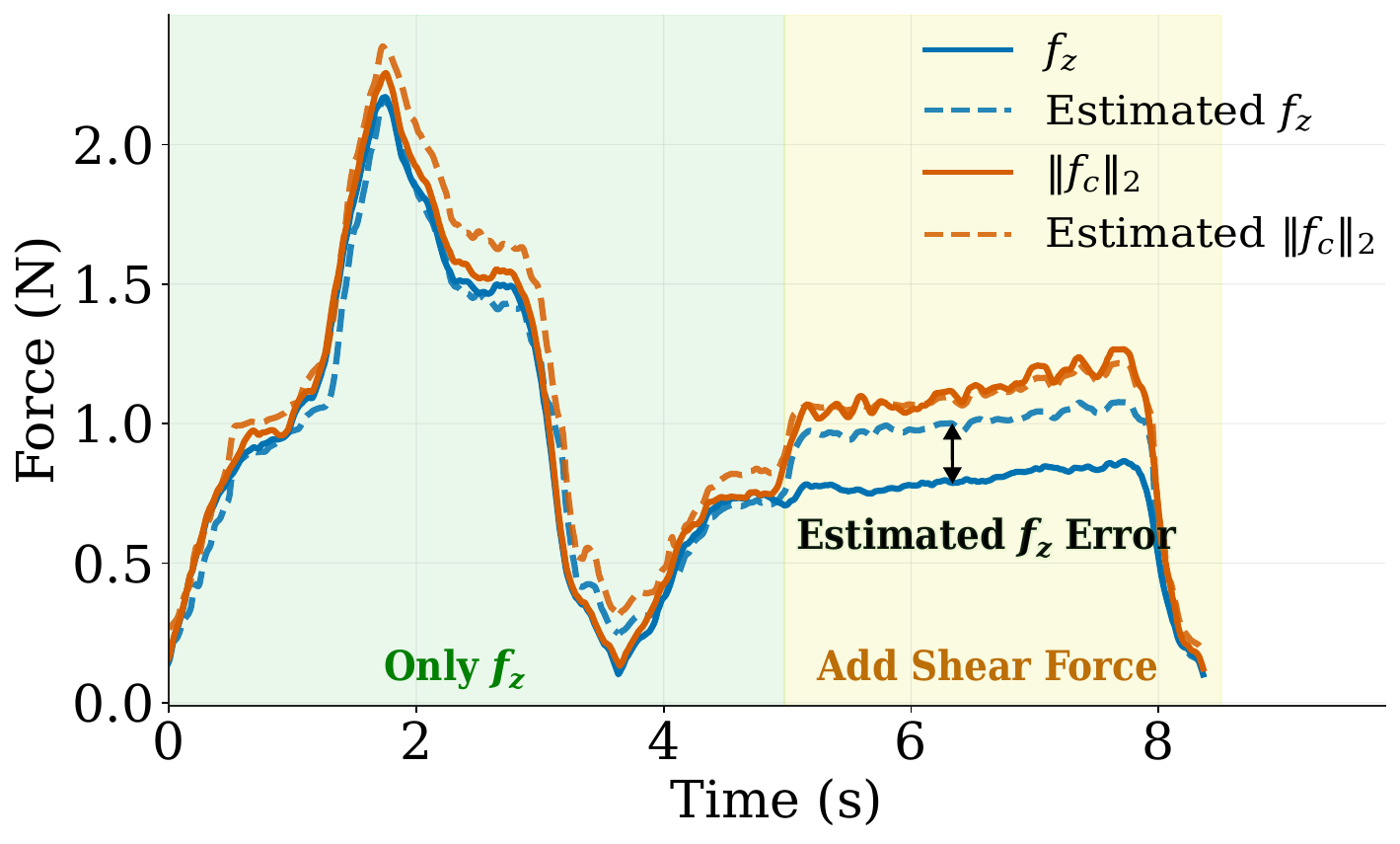}
            \put(3,4){\colorbox{white}{\textbf{(c)}}}
        \end{overpic}
    \end{minipage}

    \begin{minipage}[c]{0.55\columnwidth}
        \centering
        \begin{overpic}[width=\linewidth]{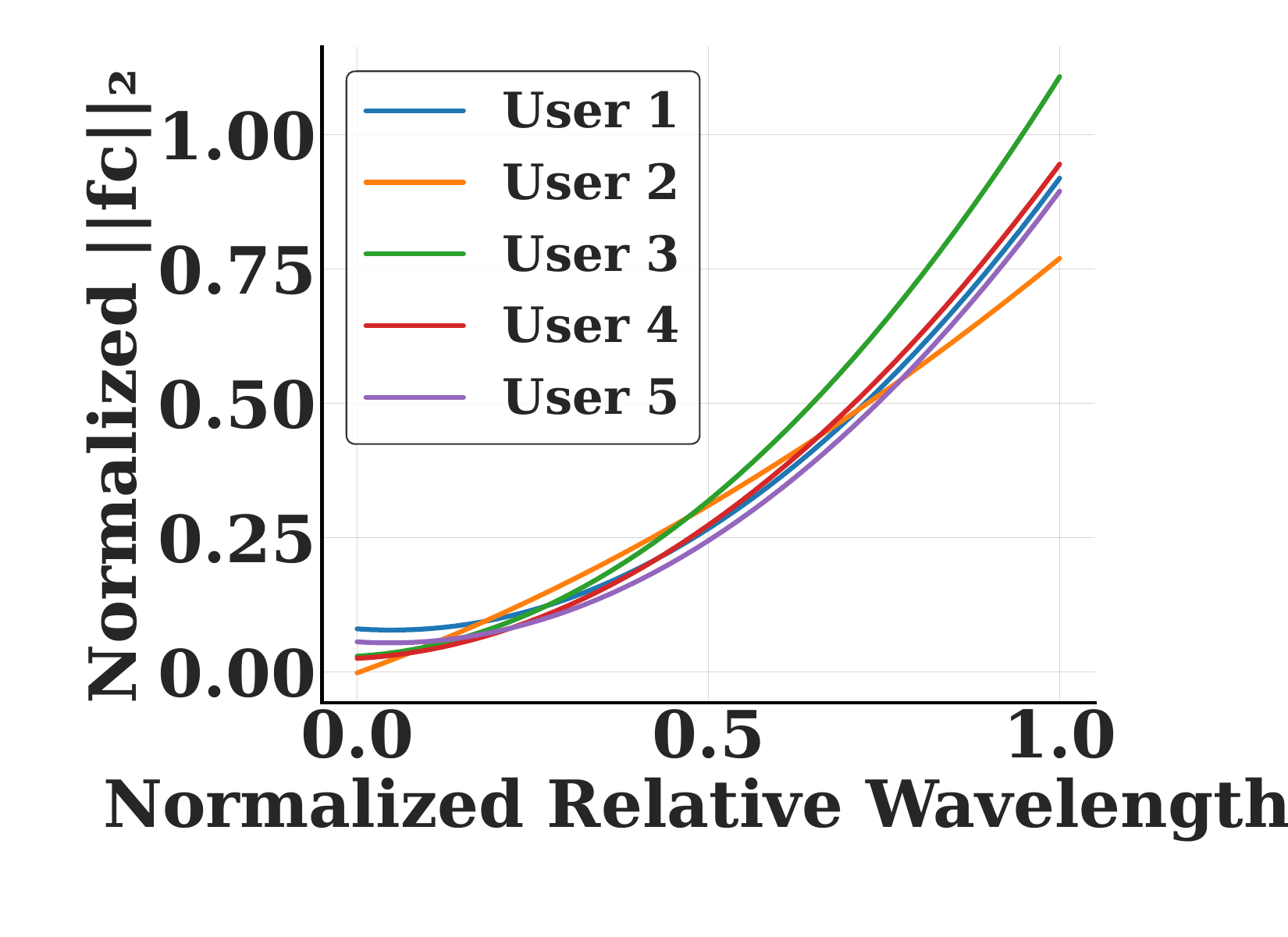}
            \put(-5,5){\colorbox{white}{\textbf{(d)}}}
        \end{overpic}
    \end{minipage}
    \begin{minipage}[c]{0.43\columnwidth}
        \centering
        \begin{overpic}[width=\linewidth]{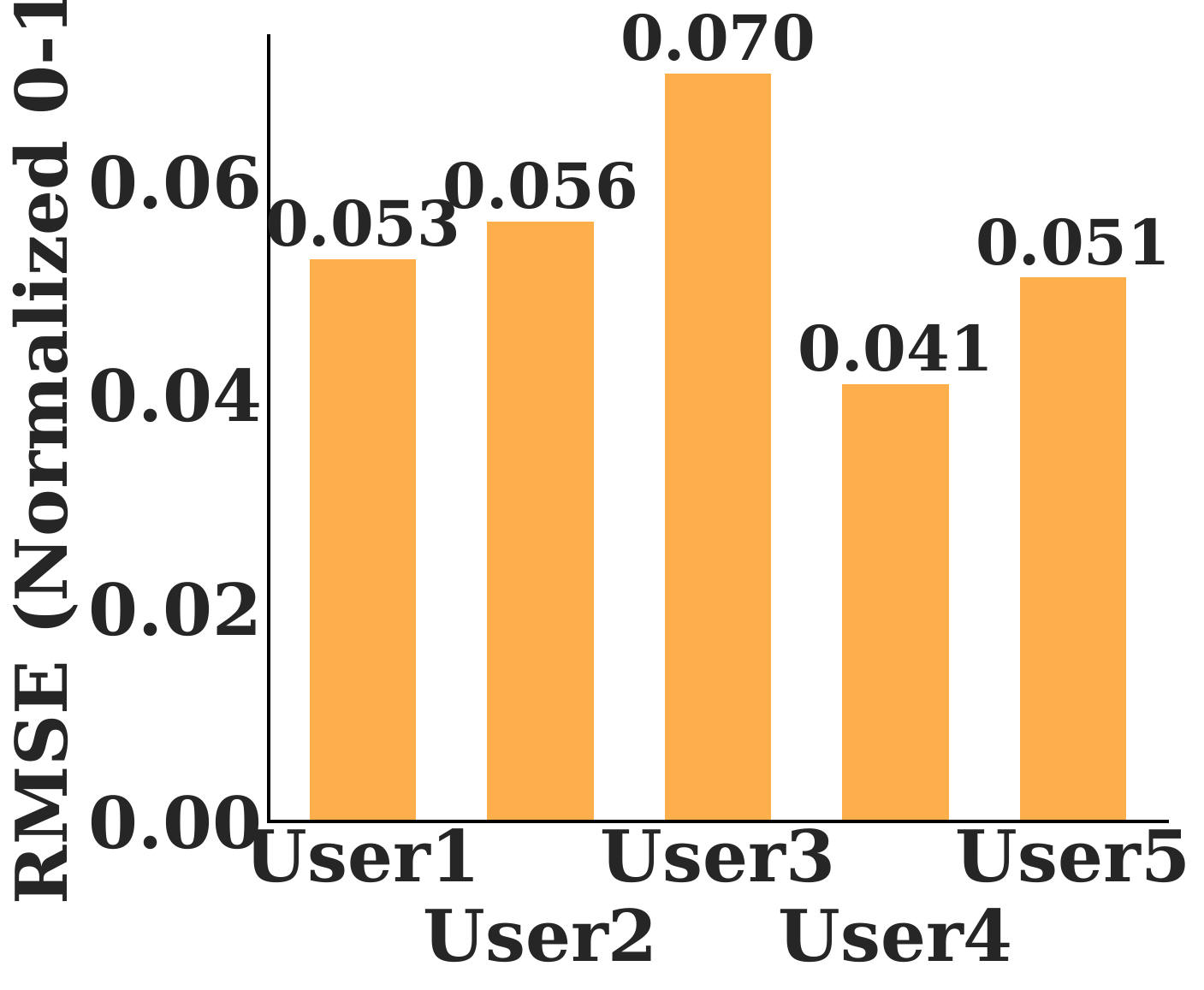}
            \put(8,0){\textbf{(e)}}
        \end{overpic}
    \end{minipage}

    \caption{
    %\mc{quasistatic? } 
    Quasi-static force-response characterization. (a) TacClip pressed against an ATI force/torque sensor. (b) Raw wavelength shift during pressing. (c) Learned estimates of $f_z$ and $\lVert \mathbf{f}_c \rVert_2$ compared with ATI measurements. In the first period (0-5\,s) the force is mostly in the normal direction; in the second period, normal and shear forces are applied, so that the estimated force is a better measure of $\lVert \mathbf{f}_c \rVert_2$ than of $f_z$. (d). Cross-user quasi-static wavelength-force curve comparison, normalized on each user. (e). Relative RMSE in $\lVert \mathbf{f}_c \rVert_2$ prediction across different users.
    % \jx{(c) is a cool plot, and we discussed it in one of our meetings. I still think (c) as it is does not convey the message clearly and strikingly enough. My understanding is that TacClip is good at estimating $\lVert \mathbf{f}_c \rVert_2$ but not forces at individual axes. I have a couple of specific suggestions. (1) Can you order the legend so that Estimated $f_z$ is right below $f_z$, so it is more obvious that these two are being compared, and the same for $f_c$? (2) In your legend, $f_c$ does not match $\lVert \mathbf{f}_c \rVert_2$. (3) Can we use a double-ended arrow to cover and highlight the gap between Estimated $f_z$ and $f_z$? (4) Instead of arrows and text, maybe we can put a vertical dashed line at "Add shear force", and a semi-transparent background with different colors to indicate 
    % ``Only $f_z$" and "Add shear force". Then you put "Only $f_z$" and "Add shear force" as normal titles for each part, without labeling with arrows. Happy to sketch out my idea on paper for you if needed. }
    }
    \label{fig:static_force}
\end{figure}

\subsection{Quasi-static Force Response}

TacClip is not a conventional force transducer. It measures contact-induced deformation of the finger--clip system, which is related to the fingertip contact force through the geometry and viscoelastic properties of the human fingertip. The main static trend is monotonic but nonlinear:
\begin{equation}
    \Delta \lambda \approx g\!\left(\left\lVert \mathbf{f}_c \right\rVert_2,\, \mathcal{H}\right),
\end{equation}
where $\Delta \lambda$ is the FBG wavelength shift, $\mathbf{f}_c=[f_x,f_y,f_z]^T$ is the contact force at the fingertip, and $\mathcal{H}$ denotes contact history effects such as tissue hysteresis and relaxation. Because the fingertip is viscoelastic, a rapid load-and-unload event can leave transient deformation after the external force has returned to zero. This behavior introduces hysteresis and motivates data-driven calibration.

We characterized the static response by wearing TacClip while pressing and sliding against an ATI force/torque sensor, which provides ground-truth $f_x$, $f_y$, and $f_z$ (Fig.~\ref{fig:static_force}(a)). Fig.~\ref{fig:static_force}(b) shows the raw wavelength shift during a contact sequence. We trained a multilayer perceptron (MLP) to map the TacClip signal and its recent history to both the normal force $f_z$ and the total force magnitude $\lVert \mathbf{f}_c \rVert_2$. Fig.~\ref{fig:static_force}(c) shows that the learned model tracks the applied load with typical error below $0.5~\mathrm{N}$ in the tested range.

The distinction between $f_z$ and $\lVert \mathbf{f}_c \rVert_2$ is important. During pure normal loading, the two quantities are identical. When shear is added while the normal component remains nearly constant, the total force magnitude increases. The TacClip wavelength shift follows the overall fingertip deformation more closely than the normal component alone, so the current single-FBG prototype is best interpreted as an estimator of contact force magnitude rather than a normal force sensor.

To evaluate generalizability, Fig. \ref{fig:static_force}(d) shows the fitted normalized quasi-static wavelength-force curves across five users, which remain highly consistent despite natural variations in finger size and initial preloading. Fig. \ref{fig:static_force}(e) presents the relative error in estimating $\lVert \mathbf{f}_c \rVert_2$ when transferring a baseline MLP model to these new users, using only a brief calibration step to account for individual tissue hysteresis. Ultimately, these results demonstrate that the TacClip system facilitates highly efficient and easily adaptable cross-user learning. 

% 5 users requires 3 sizes of different D varing from xx to xx mm. 

\begin{figure}[t]
    \centering
    \begin{minipage}[c]{0.8\columnwidth}
        \centering
        \begin{overpic}[width=\linewidth]{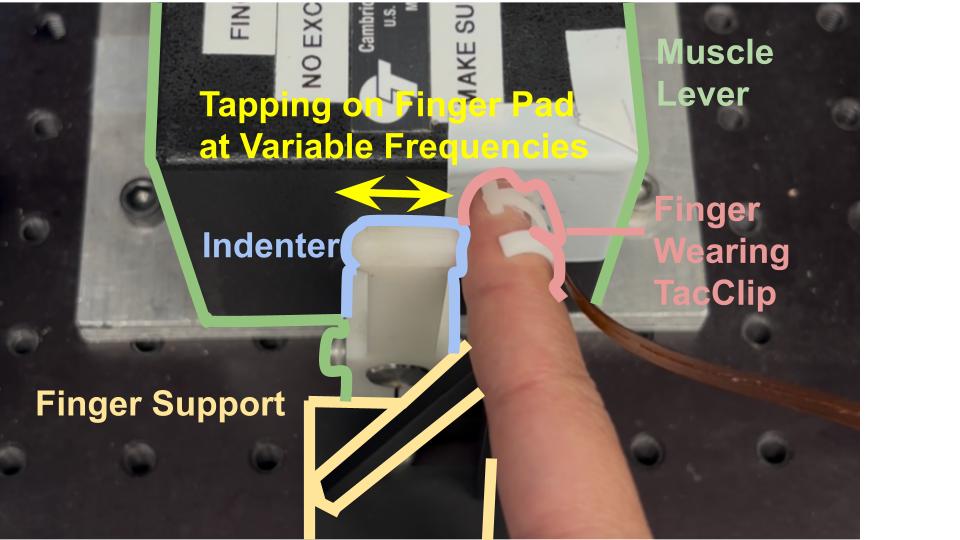}
            \put(5,5){\textcolor{white}{\textbf{(a)}}}
        \end{overpic}
    \end{minipage}
    \begin{minipage}[c]{0.9\columnwidth}
        \centering
        \begin{overpic}[width=\linewidth]{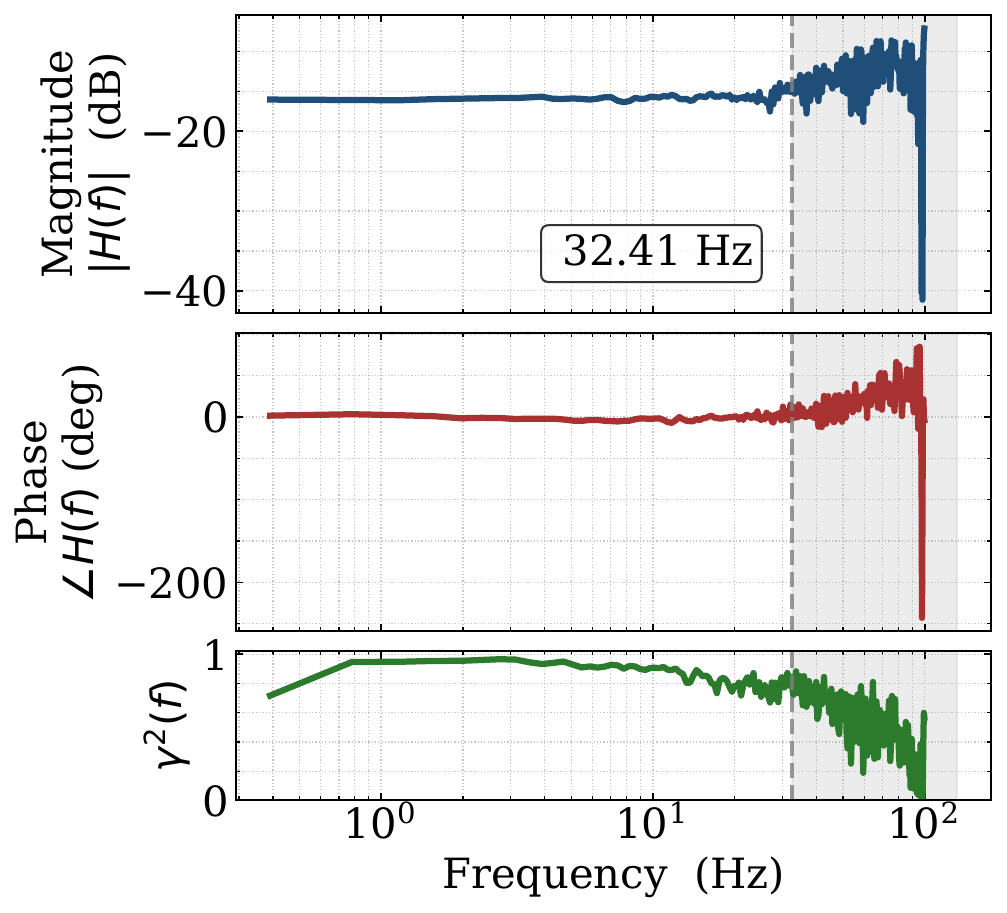}
            \put(5,5){\colorbox{white}{\textbf{(b)}}}
        \end{overpic}
    \end{minipage}

    \caption{Dynamic response under swept-frequency stimulation. (a) Muscle-lever setup applying a chirp force to the fingerpad. (b) Bode plot and coherence of the wavelength-shift response divided by the applied force.}
    \label{fig:muscle_lever_experiment}
\end{figure}

\subsection{Dynamic Tactile Sensing}

The FBG interrogator samples the TacClip signal at $2~\mathrm{kHz}$, which is sufficient to capture vibration-rich tactile events generated during sliding, tapping, and edge crossing. The limiting mechanical element is not the FBG itself but the human fingertip: soft tissue acts as a low-pass mechanical filter that attenuates high-frequency vibrations before they produce measurable clip deformation.

To characterize the frequency response under realistic wearing conditions, we used a muscle lever system (Aurora Scientific Inc., 309C) to apply an oscillatory force directly to the fingerpad while TacClip was worn. The commanded chirp had approximately constant force amplitude and a frequency increasing from $0$ to $100~\mathrm{Hz}$. Fig.~\ref{fig:muscle_lever_experiment}(a) shows the setup. Subsequently, the system transfer function from applied force to FBG wavelength shift was computed as
\begin{equation}
H(f) = \frac{\Delta\Lambda(f)}{F(f)},
\end{equation}
was computed via FFT analysis of the chirp data to quantify the mapping from the applied force to the resulting FBG wavelength shift.

As shown in Fig.~\ref{fig:muscle_lever_experiment}(b), the magnitude and phase of $H(f)$ remain relatively stable from near-DC to roughly $20$--$30~\mathrm{Hz}$, with low phase lag and high magnitude-squared coherence $\gamma^2(f)$. Above this range, coherence decreases and the magnitude and phase become less consistent, although a measurable response persists. This response is consistent with the fingertip attenuating rapid vibrations. For manipulation data collection, the result indicates that TacClip can capture the force variations associated with pressing and holding as well as a portion of the contact-induced vibration during sliding.

\begin{figure}[t]
    \centering

    \begin{minipage}[b]{0.48\columnwidth}
        \centering
        \begin{overpic}[width=\linewidth]{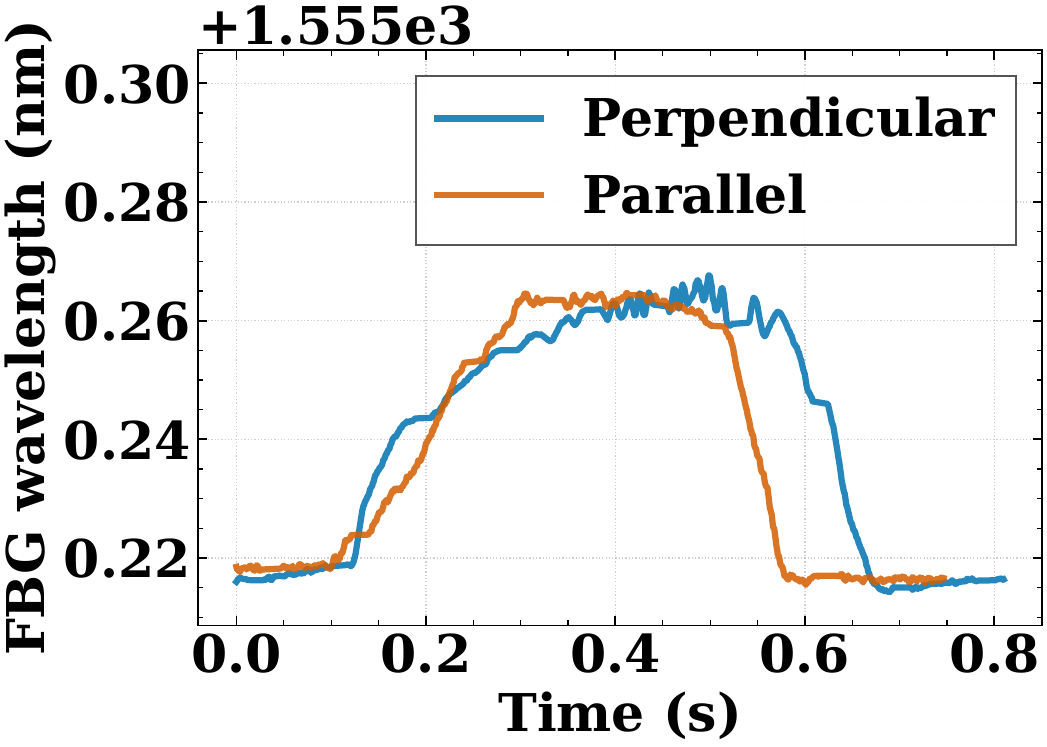}
            \put(4,4){\colorbox{white}{\textbf{(a)}}}
        \end{overpic}
    \end{minipage}
    \begin{minipage}[b]{0.48\columnwidth}
        \centering
        \begin{overpic}[width=\linewidth]{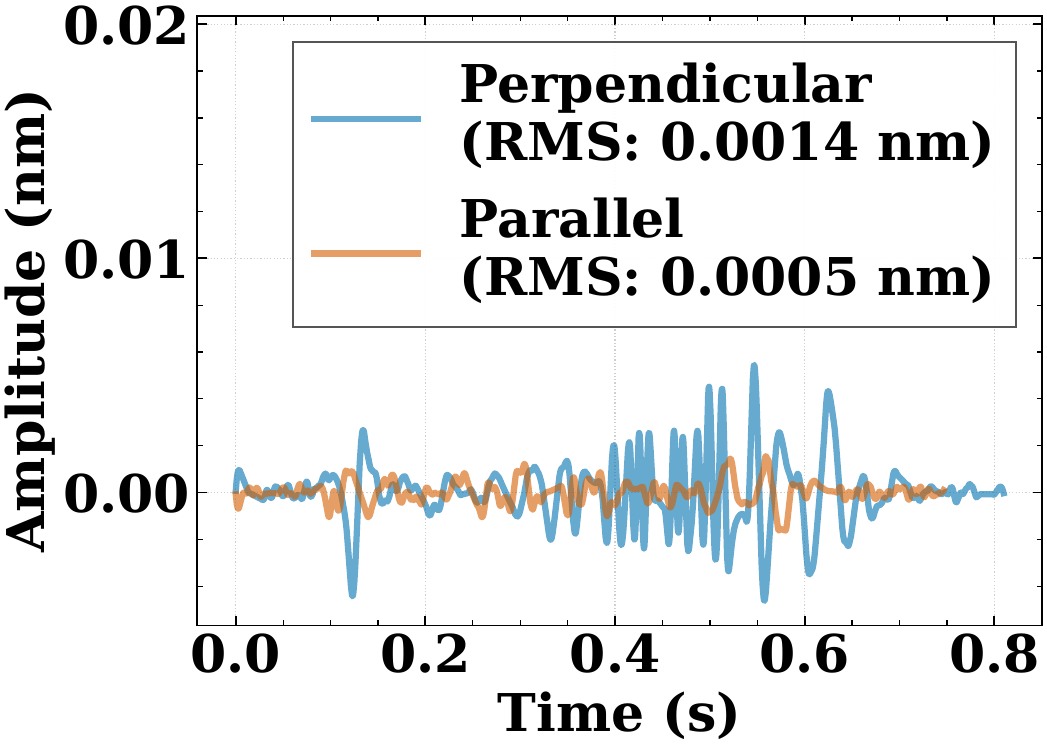}
            \put(3,4){\colorbox{white}{\textbf{(b)}}}
        \end{overpic}
    \end{minipage}
    \begin{minipage}[c]{0.39\columnwidth}
        \centering
        \begin{overpic}[width=\linewidth]{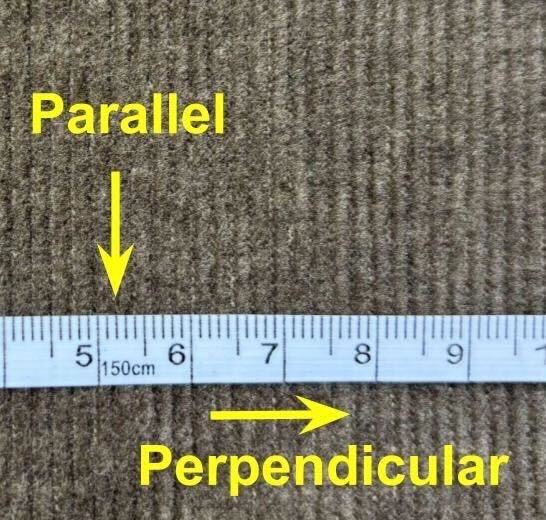}
            \put(3,8){\colorbox{white}{\textbf{(c)}}}
        \end{overpic}
    \end{minipage}
    % \hfill
    \begin{minipage}[c]{0.58\columnwidth}
        \centering
        \begin{overpic}[width=\linewidth]{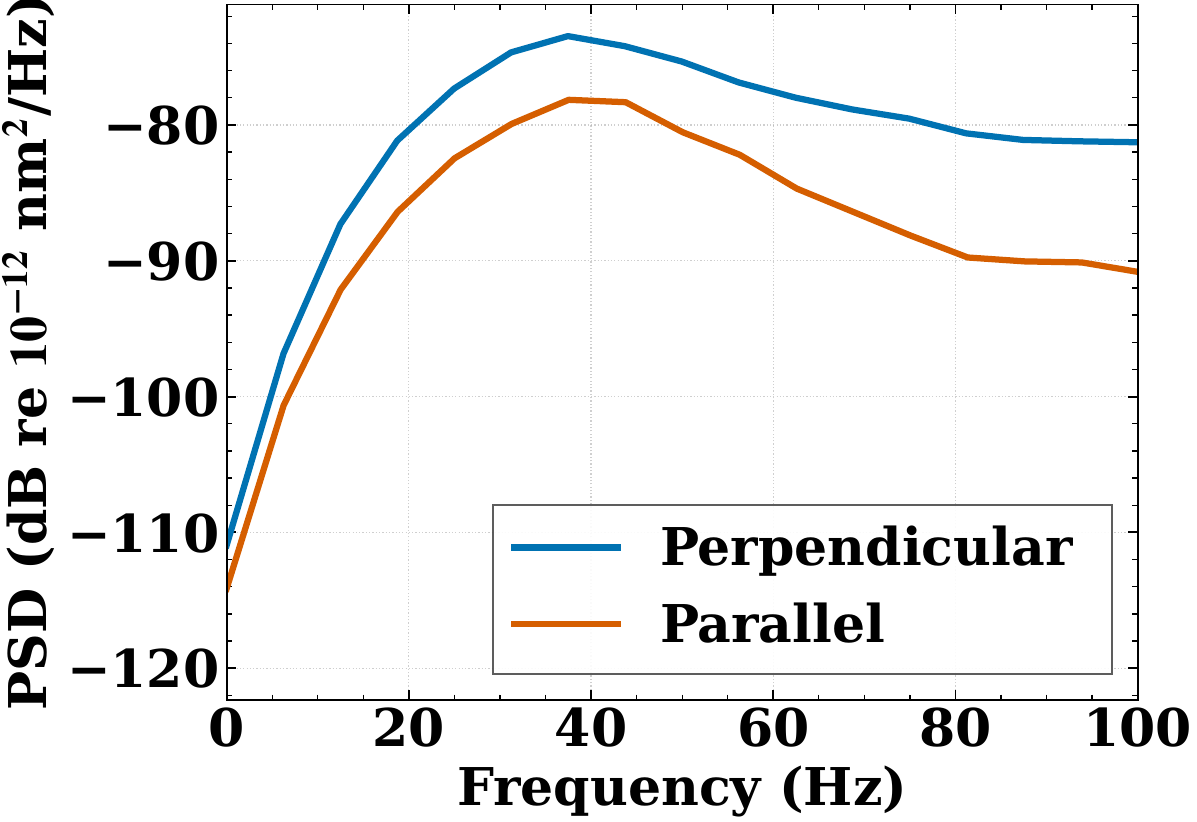}
            \put(4,4){\colorbox{white}{\textbf{(d)}}}
        \end{overpic}
    \end{minipage}
    \caption{Directional texture sensing on corduroy fabric. (a) Raw TacClip sensor signals captured during sliding motions perpendicular and parallel to the fabric ridges. (b) High-pass filtered signals ($>30\text{ Hz}$) highlight distinct tactile signatures; sliding perpendicular to the ridges exhibits more regular and intense periodic vibrations compared to parallel sliding. (c) Photograph of the corduroy test sample. (d) Power spectral density (PSD) of the dynamic wavelength signals for both sliding directions. 
    %\jx{fig (a) and (b) have some occlusions. Also, I always recommend PDF over PNG. PDF is vectorized. }
    %\mc{Plots (a) and (b) are small and a bit hard to read. What do we want the reader to clearly see from them?} 
%    \mc{Can we make corduroy photo brighter and add a scale bar?}
%MRC This looks great!
%    \ivy{Updated! Both figures are now fixed and replaced with PDFs.}
    }
    \label{fig:cloth_align}
\end{figure}

\begin{figure*}[t]
    \centering
    \begin{subfigure}[c]{0.8\columnwidth}
        \centering
        \begin{overpic}[width=\linewidth]{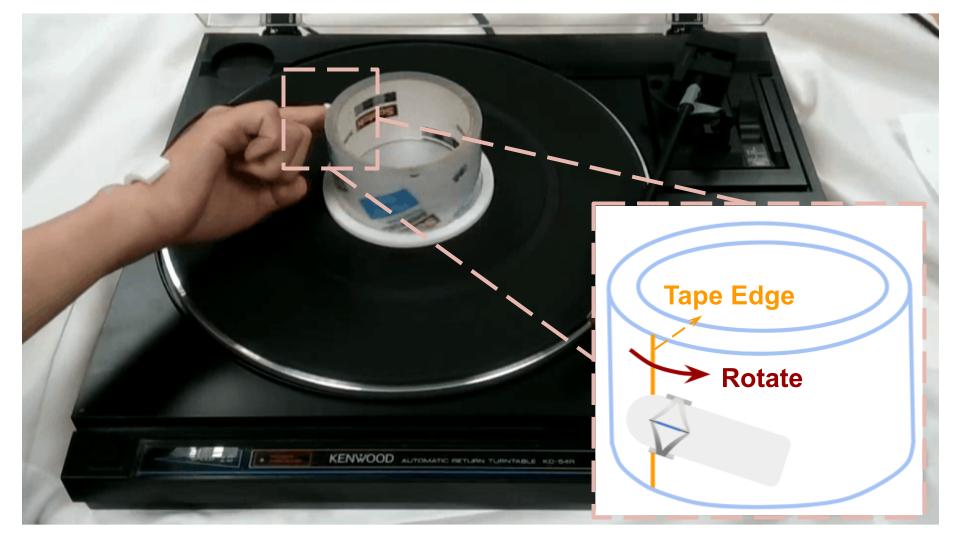}
            \put(5,5){\textcolor{white}{\textbf{(a)}}}
        \end{overpic}
    \end{subfigure}
    \begin{subfigure}[c]{1.1\columnwidth}
        \centering
        \begin{overpic}[width=\linewidth]{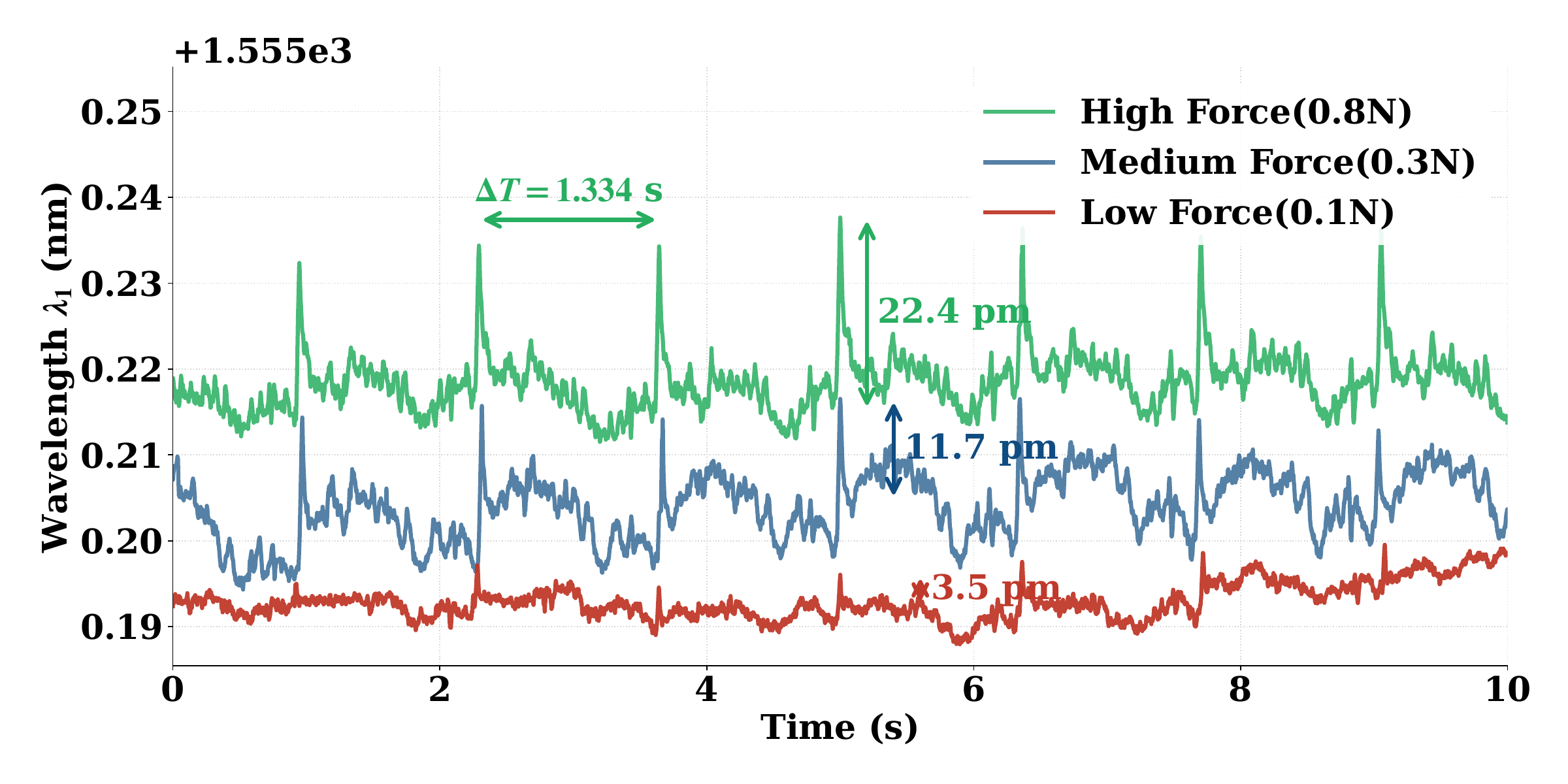}
            \put(5,5){\colorbox{white}{\textbf{(b)}}}
        \end{overpic}
    \end{subfigure}
    
    \caption{Transparent tape-edge detection. (a) Turntable setup for repeatable sliding over a transparent tape edge. (b) TacClip wavelength signal for three pressing forces and time interval between detection peaks using high pressing force.}
    
    %MC I would like to show the standard time between peaks (pick one pair) for the "high" force case with horizontal double arrow showing time in ms. Remember, the turntable rpm is not what is important. The linear surface speed r*\omega is what is important. 
    
   % Is boundary detect error \pm (plus minus)?  and is it measured as standard deviation? As standard error?
   
   %Also how accurately is the force measure. When you say the force is low (0.1 N) what is the variation or uncertainty in that number?

   %Ivy Answer: the resolution of TacClip is almost 0.1N, so it's the smallest force that it could sense. But it's hard for finger to maintain such a small force, so the actual data usually sits a bit above that. But the output signal is stable even baseline force is unstable.
    
    \label{fig:record_player_experiment}
\end{figure*}

\begin{table*}[t]
    \centering
    \caption{Tape edge detection under different pressing forces 
%    \jx{Should there be a space between ``Peak Interval'' and ``(s)''. Also, you should be consistent whether the table/figure title is capitalized. I am a bit confused by this table of results. When I see something like ``Detection Restults'', I think of success rate. Why is here the unit ``second''? I guess you can make it more specific here in the table caption that the metric here is the detected time of the edge vs the ground truth time of the edge passing the finger tip?} 
% \mc{See comments in the source.}
 %MRC This table tells me more than I really care to know about detected tape edge timing. I think I only care how reliably the tape edge is detected and with what spatial accurcy. So, from that standpoint I would ask for:
% Detected interval\\ (1.333 s ground truth)
% mean interval timing error, s (and possibly also std deviation, s)
% mean edge location error, mm (and possibly also std deviation, mm)
%But really, just the mean error in timing and edge location is probably enough
%Also, at the lightest pressure, did it ever fail to detect the edge? 
    }
    \label{tab:tape_edge_results}
    \begin{tabular}{l c c c c c c}
        \toprule
        \textbf{Force Level} & \makecell{Detected Interval \\ Ground Truth (s)} & \makecell{Detected Interval (s)} & \makecell{Mean Interval \\ Timing Error (s)} & \makecell{Interval \\ Timing Std. (s)} & \makecell{Mean Edge \\ Location Error (mm)} & \makecell{Edge Location \\ Std. (mm)} \\
        \midrule
        Low ($0.1~\mathrm{N}$, n=23)  &\multirow{3}{*}{1.333} & 1.384 & +0.051 & 0.340 & +10.7  & 15.3 \\
        Medium ($0.3~\mathrm{N}$, n=27) & & 1.321 & -0.012 & 0.104 & -2.7 & 4.7 \\
        High ($0.8~\mathrm{N}$, n=26) & & 1.338 & +0.005 & 0.013 & +1.0 & 0.6 \\
        \bottomrule
    \end{tabular}
\end{table*}

\section{Application Experiments}
\label{sec:experiments}

The preceding characterization shows that TacClip can measure both contact-force magnitude trends and dynamic tactile events. We next evaluate tasks in which bare-finger tactile cues are useful and where vision alone can be ambiguous: sensing fabric direction, detecting a transparent tape edge, and measuring contact forces underwater.

\subsection{Cloth Alignment}

Many fabrics have fine-scale directional features, including ribs, seams, or weave patterns. Humans can use lateral exploratory motion to determine these directions before aligning or folding cloth. We tested whether TacClip can detect directional differences by sliding a fingertip over corduroy either perpendicular or parallel to the ridges.

Fig.~\ref{fig:cloth_align}(a) compares the raw TacClip signals recorded during a single sliding motion both perpendicular and parallel to the fabric ridges. Fig.~\ref{fig:cloth_align}(b) illustrates these signals after applying a $30~\mathrm{Hz}$ high-pass filter to remove the effects of the baseline contact force. Under comparable applied forces, sliding perpendicular to the ridges induces significantly larger vibration amplitudes (approximately $\pm 0.004~\mathrm{nm}$ wavelength variations) than sliding parallel to the ridges (approximately $\pm 0.001~\mathrm{nm}$). The power spectral density analysis in Fig.~\ref{fig:cloth_align}(d) further distinguishes these two conditions within the frequency domain. Together, these results demonstrate that TacClip effectively captures direction-dependent texture cues while preserving the user's bare-finger perception of the cloth.

\subsection{Transparent Tape Edge Detection}

Humans commonly use their fingertips or fingernails to find the edge of a roll of transparent tape. The edges are often difficult to localize visually because their appearance depends strongly on specular reflection, illumination, and background texture. A fingertip, however, can detect the small surface-height discontinuity and friction change at the edge. 

%MRC We should say the tape thickness. I'm guessing we used 3M "heavy duty packing tape" whic is 3 mil = 
We evaluated TacClip on this task by placing a roll of $78\,\mu\mathrm{m}$ thick packing tape on a turntable to produce a repeatable sliding motion under the user's stationary finger (Fig.~\ref{fig:record_player_experiment}(a)). 
The turntable rotated at $45~\mathrm{RPM}$, corresponding to 
%an angular velocity of $\omega \approx 4.71~\mathrm{rad/s}$ and 
a surface velocity of $21.25~\mathrm{cm/s}$, slightly higher than typical velocities used by humans in stereotypical exploratory procedures for detecting fine surface features 
\cite{exploration_procedure}. 
% This is because human tactile perception of fine textures typically favors slower speeds (8–10 cm/s), whereas the detection of tape boundaries is essentially an edge detection task which could be preformed under faster speed. 

Experiments were conducted under three pressing force levels: $0.1\,\mathrm{N}$, $0.3\,\mathrm{N}$, and $0.8\,\mathrm{N}$. As shown in Fig.~\ref{fig:record_player_experiment}(b), repeatable peaks are observed when the fingertip traverses the tape boundary. Increasing the pressing force enhances both the peak amplitude and the signal-to-noise ratio (SNR). To reliably detect these peaks, we employed an adaptive thresholding strategy. At lower force levels, the data lacks distinct peaks, allowing for a lower detection threshold; conversely, at higher force levels, the threshold is adjusted upward to ensure robust peak identification. By leveraging these adaptive thresholds, we successfully captured all boundary-crossing events. Table~\ref{tab:tape_edge_results} summarizes the experimental results. Given the tape’s angular velocity, the theoretical time interval between consecutive peaks is $1.33\,\mathrm{s}$. Using this as a baseline, we calculated the mean and standard deviation of the boundary detection error. The results demonstrate that as pressing force increases, both the accuracy and precision of the detection improve, and the range between the maximum and minimum interval windows narrows. This improvement is attributed to the higher SNR achieved at greater contact forces.
% ; at lower forces, the signal is insufficient, leading to a higher susceptibility to missed or erroneous detections. 
These findings are consistent with the principle that stronger contact produces more distinct vibrations when traversing the tape edge.

\subsection{Underwater Force Response}

Unlike many electronic transduction methods, optical fibers and 
FBG sensors are inherently waterproof and unaffected even by saltwater. This property is attractive for collecting human tactile demonstrations in wet applications ranging from washing to marine applications. To confirm that TacClip is not affected by immersion in water we performed a spring compression test in air and in water.

\begin{figure}[htbp]
    \centering
    \begin{minipage}[c]{0.32\columnwidth}
        \centering
        \begin{overpic}[width=\linewidth, trim=0 2cm 0 4cm, clip]{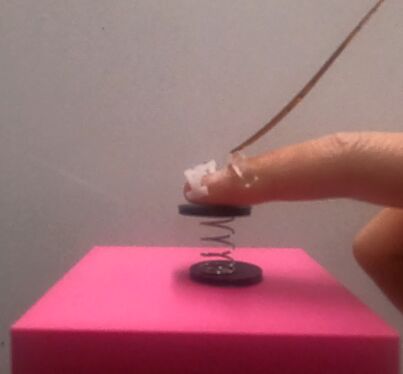}
            \put(5,5){\colorbox{white}{\textbf{(a)}}}
        \end{overpic}
    \end{minipage}
    \hfill
    \begin{minipage}[c]{0.32\columnwidth}
        \centering
        \begin{overpic}[width=\linewidth, trim=0 2cm 0 4cm, clip]{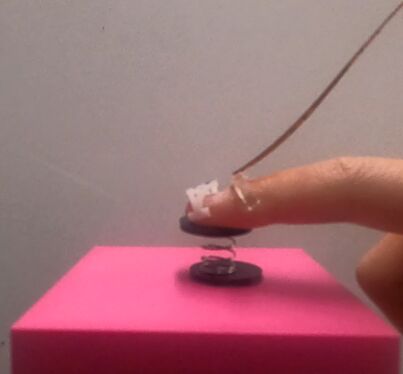}
            \put(5,5){\colorbox{white}{\textbf{(b)}}}
        \end{overpic}
    \end{minipage}
    \hfill
    \begin{minipage}[c]{0.32\columnwidth}
        \centering
        \begin{overpic}[width=\linewidth, trim=0 2cm 0 4cm, clip]{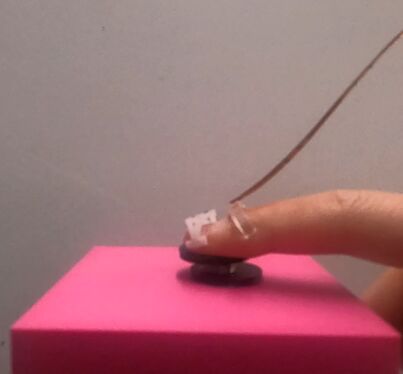}
            \put(5,5){\colorbox{white}{\textbf{(c)}}}
        \end{overpic}
    \end{minipage}
    \begin{minipage}[c]{0.32\columnwidth}
        \centering
        \begin{overpic}[trim=0 2cm 0 2cm, clip, width=\linewidth, angle=180]{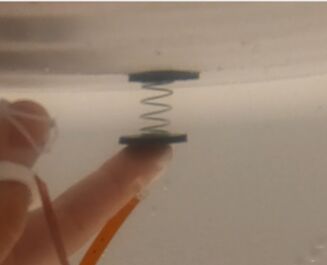}
            \put(5,5){\colorbox{white}{\textbf{(d)}}}
        \end{overpic}
    \end{minipage}
    \hfill
    \begin{minipage}[c]{0.32\columnwidth}
        \centering
        \begin{overpic}[trim=0 2cm 0 2cm, clip, width=\linewidth, angle=180]{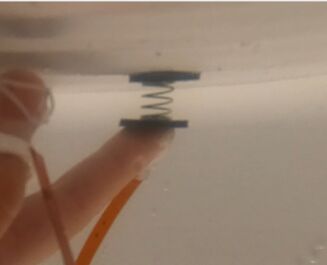}
            \put(5,5){\colorbox{white}{\textbf{(e)}}}
        \end{overpic}
    \end{minipage}
    \hfill
    \begin{minipage}[c]{0.32\columnwidth}
        \centering
        \begin{overpic}[trim=0 3cm 0 1cm, clip, width=\linewidth, angle=180]{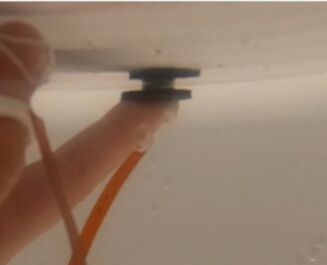}
            \put(5,5){\colorbox{white}{\textbf{(f)}}}
        \end{overpic}
    \end{minipage}

    \vspace{0.5cm}
    
    \begin{minipage}[c]{0.95\columnwidth}
        \centering
        \begin{overpic}[width=\linewidth]{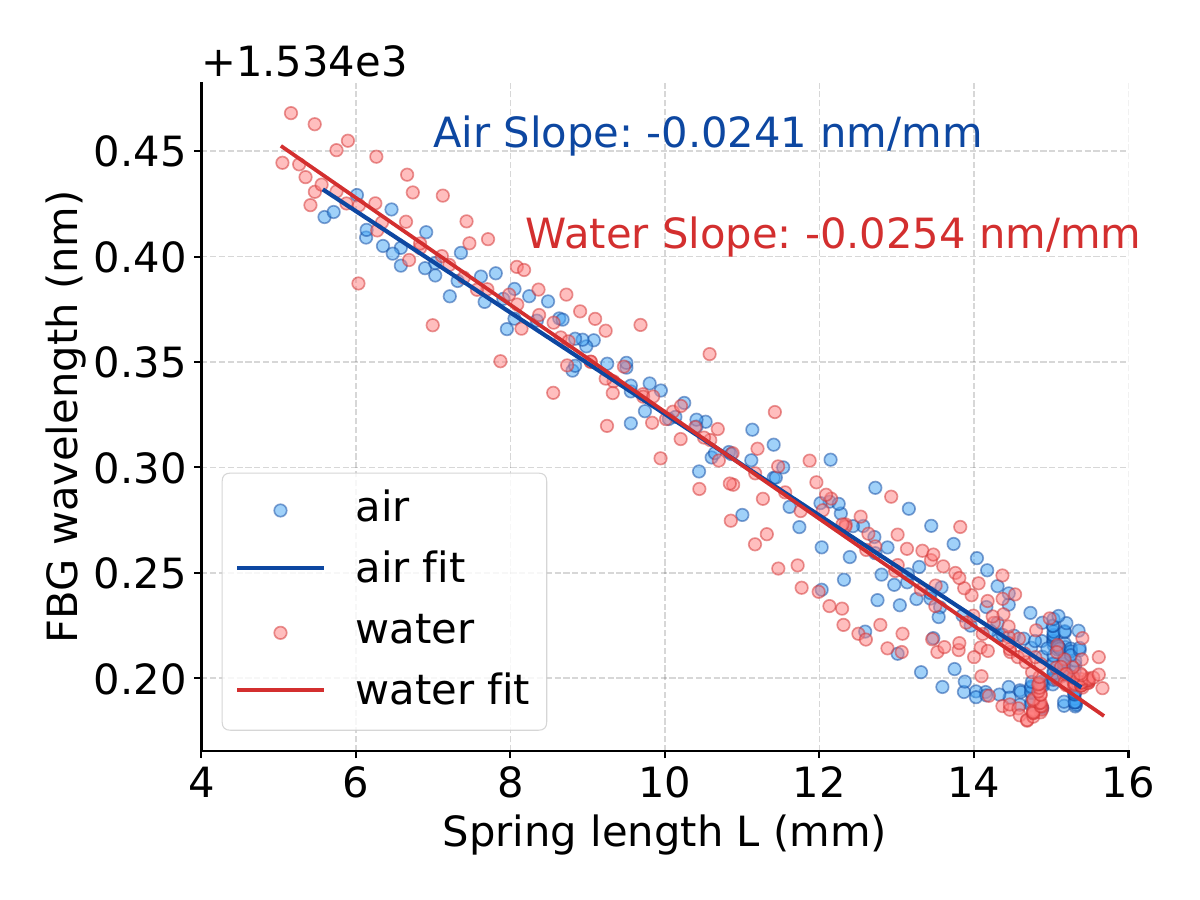}
            \put(5,5){\colorbox{white}{\textbf{(g)}}}
        \end{overpic}
    \end{minipage}

    \caption{Underwater spring-compression response. (a)--(c) Compression sequence in air. (d)--(f) Compression sequence underwater. (g) TacClip wavelength shift versus spring length in air and water after temperature calibration. Similar slopes indicate similar mechanical sensitivity.
   %If we subtract a baseline zero force reading, is that not sufficient to calibrate for air versus water?
    }
    \label{fig:spring}
\end{figure}

A compression spring with flat plates bonded to both ends was depressed by a human fingertip while TacClip recorded wavelength shifts. The spring length, measured using a camera recording video from the side, served as a proxy for applied force. The estimated spring length measurement accuracy is approximately $\pm 1.0$\,mm for each reading. 
%\mc{The estimated spring length measurement accuracy is approximately $\pm X.X$\,mm for each reading.} 
% In the video, 1 mm corresponds to 40 pixels. Based on 20 manual annotations of the spring end, std is 46 pixels, measurement accuracy is approximately $\pm 1.0$ mm. 
Fig.~\ref{fig:spring}(a)--(c) and Fig.~\ref{fig:spring}(d)--(f) show representative compression sequences in air and water, respectively. Fig.~\ref{fig:spring}(g) plots wavelength shift versus spring length. After temperature calibration, the air and water curves are not significantly different. This result suggests that TacClip can be used for wet or underwater tactile data collection if temperature is compensated.

\subsection{Surgical Palpation}
\label{sec:palpation}

%\jx{Starting from here, it does read a bit more rushed. There are also a few errors here and there. Maybe proofread and wordsmith these paragraphs a bit more before submission.}
Medical palpation relies on perceiving tactile cues for tasks
such as locating underlying blood vessels or detecting subcutaneous masses to guide subsequent procedures like injections~\cite{McKinley2015ASH}. In contrast to some data gloves, TacClip meets these requirements by providing users with unobstructed tactile sensitivity to detect changes in compliance and texture.

%\jx{Try being consistent, whether you use Fig. or Figure?} 
% Fig. \ref{fig:surgical_palpation} demonstrates an application using clinical palpation exemplars. Fig. \ref{fig:surgical_palpation}(a) shows a finger wearing the TacClip device sliding across a vascular area of the tissue phantom. In Fig.~\ref{fig:surgical_palpation}(b) there are evident dynamic signals superimposed on a lower-frequency signal that corresponds to each sliding motion. Fig. \ref{fig:surgical_palpation}(d) shows the corresponding signal for sliding across a non-vascular area of the phantom; only the gentler low-frequency variation associated with each sliding motion is observable. 

\begin{figure}[t]
     \centering
     \begin{subfigure}[c]{0.38\columnwidth}
         \centering
         \includegraphics[width=\linewidth]{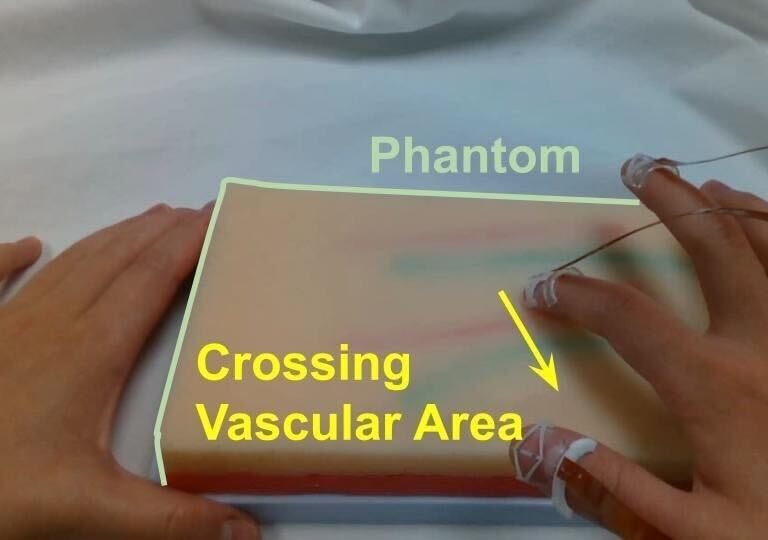}
            \put(-92,5){\colorbox{white}{\textbf{(a)}}}
     \end{subfigure}
     \begin{subfigure}[c]{0.53\columnwidth}
         \centering
         \includegraphics[width=\linewidth]{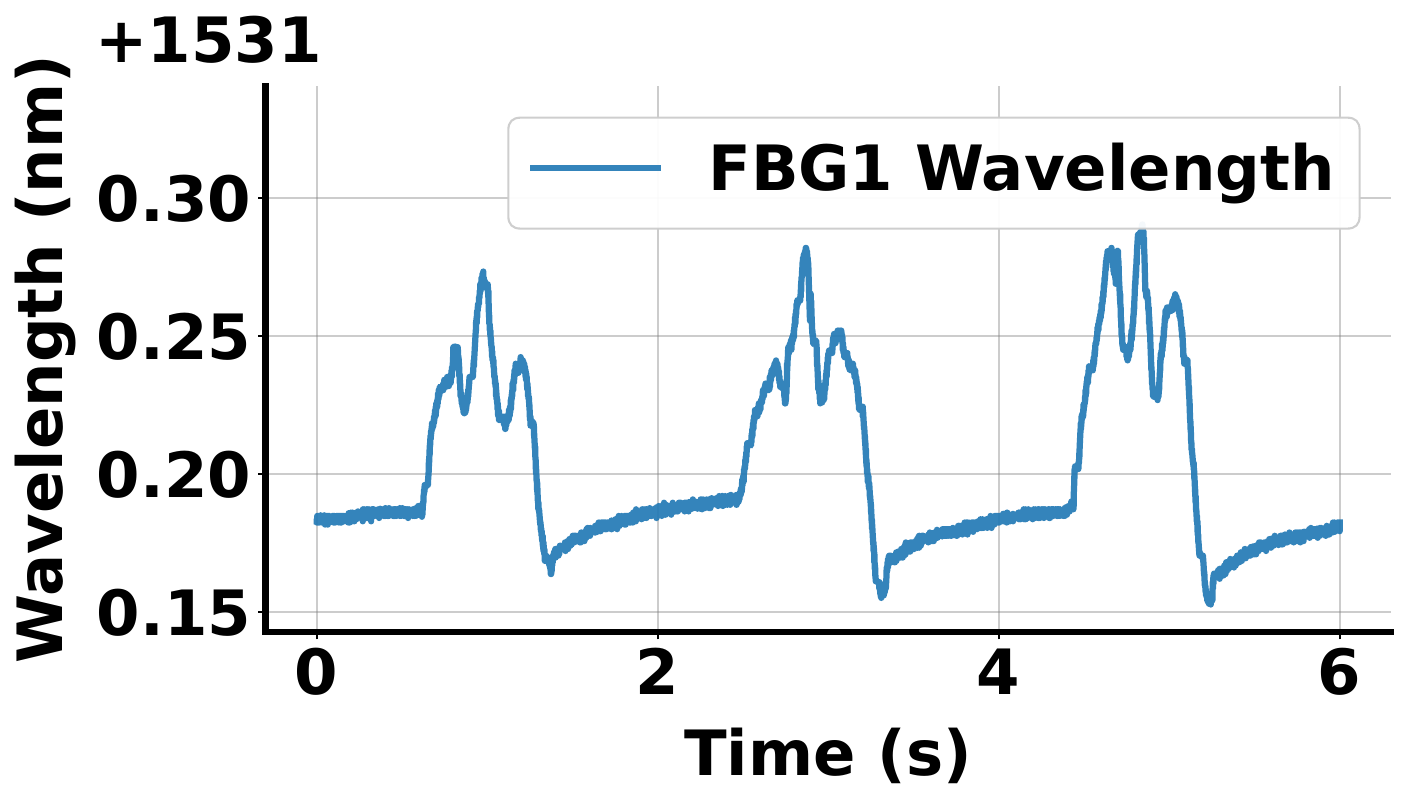}
             \put(-110,10){\colorbox{white}{\textbf{(b)}}}
     \end{subfigure}
     
     \begin{subfigure}[c]{0.38\columnwidth}
         \centering
         \includegraphics[width=\linewidth]{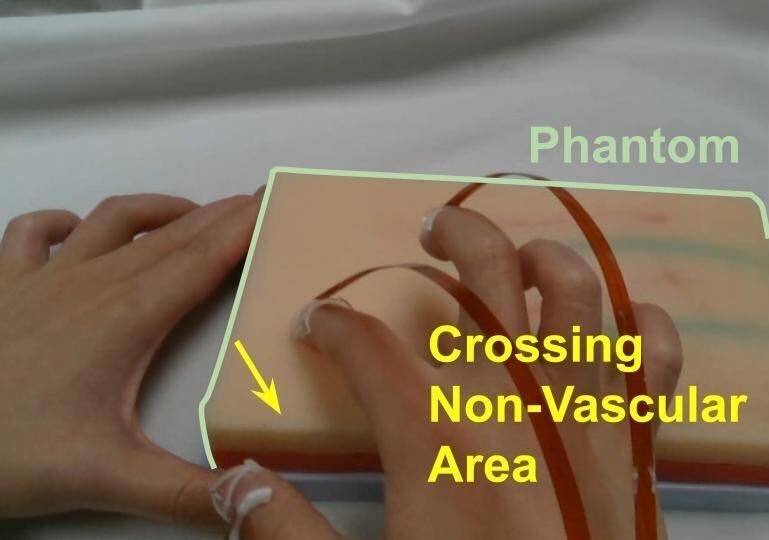}
            \put(-92,5){\colorbox{white}{\textbf{(c)}}}
     \end{subfigure}
     \begin{subfigure}[c]{0.53\columnwidth}
         \centering
         \includegraphics[width=\linewidth]{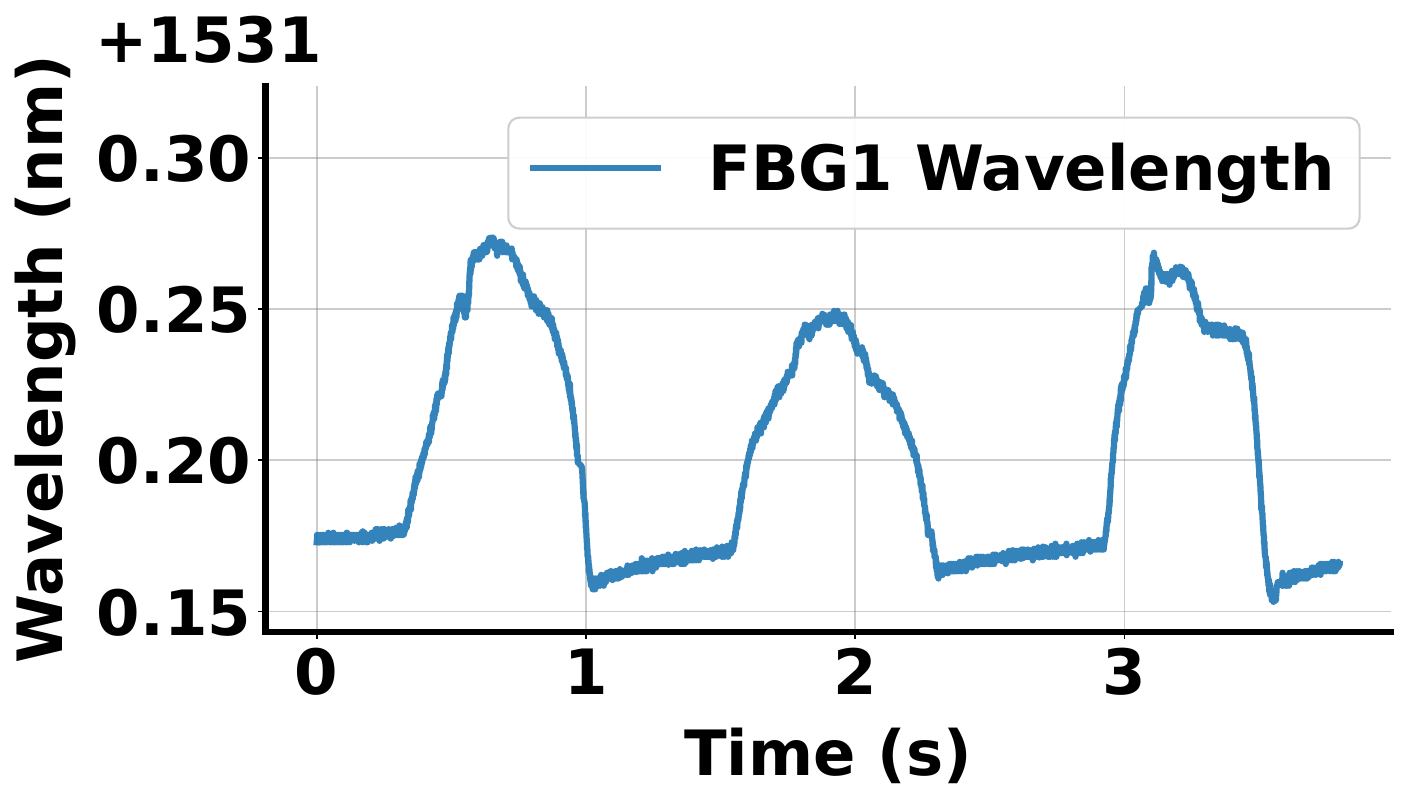}
            \put(-110,10){\colorbox{white}{\textbf{(d)}}}
     \end{subfigure}
     
     \caption{(a) Demonstration of sliding across a vascular area of a tissue phantom. (b) TacClip signals acquired during the process. (c) Demonstration of sliding across a non-vascular area. (d) TacClip signals acquired during the process. 
     }
     \label{fig:surgical_palpation}
\end{figure}

To evaluate this capability, we conducted a palpation experiment on a tissue phantom (Fig.~\ref{fig:surgical_palpation}(a)). When sliding the TacClip-instrumented finger across a vascular region, high-frequency dynamic variations are superimposed on the baseline low-frequency motion profile (Fig.~\ref{fig:surgical_palpation}(b)). For comparison, traversing a non-vascular region produces only smooth, low-frequency signal variations resulting from the sliding motion itself (Fig.~\ref{fig:surgical_palpation}(d)). This spectral disparity 
shows that TacClip can provide dynamic signals associated with subcutaneous features during palpation.

\begin{figure*}[t]
    \centering
    \begin{subfigure}[b]{0.9\textwidth}
         \centering
         \includegraphics[trim=0pt 200pt 0pt 0pt, clip=true, width=\textwidth]{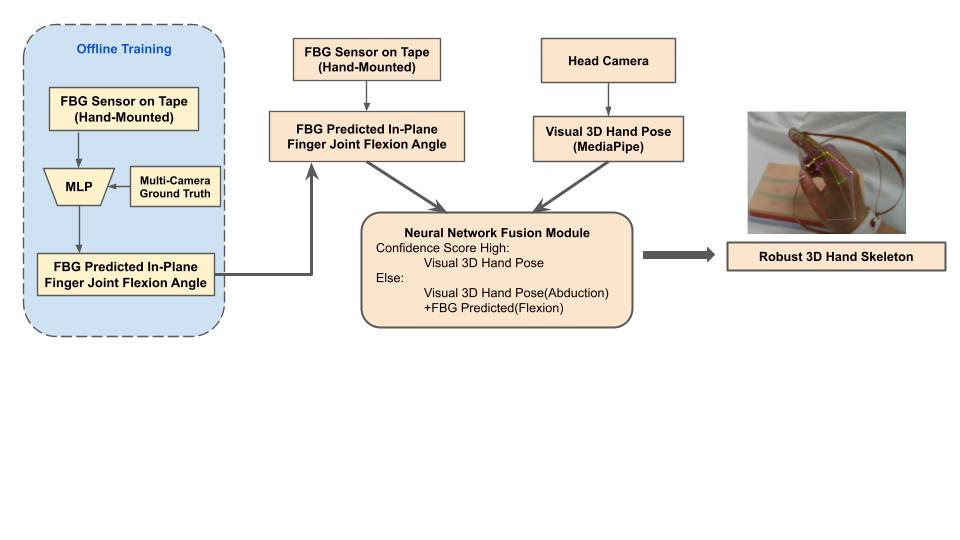}
     \end{subfigure}
     
     \caption{Schematic of the proposed vision-FBG fusion framework for robust hand skeleton tracking. 
     % \mc{Say or show more about camera setup for fusion. Also I think we need some results. One possibility is to grasp blocks (small/large), cylinders (small/large), sphere, card  -- and show the dimension error with (i) vision only 'V' and (ii) vision and TacClip 'V+T'. Hopefully we see improvement. It seems like the 'ground truth' MoCap is not actually super accurate.}
     }
     \label{fig:camera_fusion}
\end{figure*}

\begin{table*}[htbp]
\centering
\caption{Comparison of hand pose estimation accuracy during object grasping. 
%\jx{Capitalization or not?}\ivy{changed title to capitalize only on first letter}
}
\label{tab:tracking_comparison}
\begin{tabular}{lcccccc}
\toprule
\multicolumn{1}{c}{Shape} & 
\multicolumn{2}{c}{\makecell{Cylinder (n=20) \\ \includegraphics[angle=270, width=1.5cm]{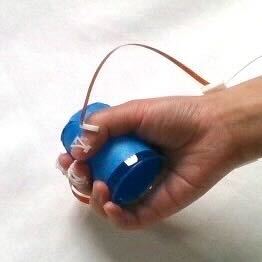}}} & 
\multicolumn{2}{c}{\makecell{Box (n=23) \\ \includegraphics[angle=270, width=1.5cm]{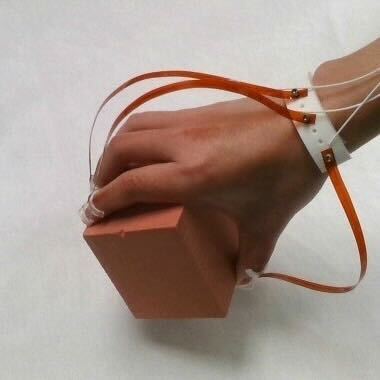}}} & 
\multicolumn{2}{c}{\makecell{Card (n=15) \\ \includegraphics[angle=270, width=1.5cm]{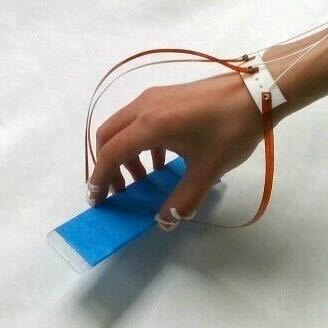}}} \\ 
\cmidrule(lr){2-3} \cmidrule(lr){4-5} \cmidrule(lr){6-7}

\multicolumn{1}{c}{Tracking Method} & 
Vision only & Vision+FBG & Vision only & Vision+FBG & Vision only & Vision+FBG \\ 
\midrule

\multicolumn{1}{c}{Reference Dimension} & \multicolumn{2}{c}{Cylinder Diameter (4.23cm)} & \multicolumn{2}{c}{Box Side Diagonal (8.97cm)} & \multicolumn{2}{c}{Card Width (3.47cm)} \\ 
\midrule

\multicolumn{1}{c}{Estimated Distance} & 
\multicolumn{2}{c}{\makecell{Thumb-Middle \\ Grasp Diameter}} & 
\multicolumn{2}{c}{\makecell{Euclidean Distance between \\ Middle Fingertip and Thumb}} & 
\multicolumn{2}{c}{\makecell{Euclidean Distance between \\ Middle Fingertip and Thumb}} \\ 
\midrule

\multicolumn{1}{c}{Mean (cm)} &  4.69cm &  4.32cm  & 10.29cm & 9.11cm & 3.91cm & 3.53cm  \\ 
\multicolumn{1}{c}{Std (cm)}  &  1.30cm &  0.74cm  & 2.13cm & 0.83cm  & 1.06cm & 0.53cm \\ 
\bottomrule
\end{tabular}
\end{table*}

\subsection{Augmenting Hand Pose Information}
\label{sec:augmenting}

As noted earlier, the same optical fiber that is used to measure fingertip deformation can also be bonded to a flexible polyimide (Kapton) strip that extends across the back of the hand to a wristband. By locating FBG sensors at regular intervals along this strip, the in-plane flexion of the fingers can be estimated by measuring the bending of the strip at multiple locations. 
Although this measurement does not fully capture finger pose, it can provide useful information to augment hand pose estimation from vision, especially when the fingers are partially occluded.

To demonstrate this capability, we applied a multimodal fusion framework that combines a head mounted camera's 3D hand pose estimation with the FBG strain data via a neural network. We evaluated the framework for grasping with the thumb, index, and middle fingers.
%MRC Is this right? Or were there fewer FBGs? Maybe just one? 

%Ivy There are 4 FBGs sensing bending in total, distrbuted along tape with regular interval. Since the tape is only fixed at the fingertip and the wristband, the boundary conditions keep changing as the finger bends. The tape's shape gets a lot more complex than its initial state, local curvature peak shifts. Because of this, the strain isn't just monotonically increasing for the middle FBGs. So, the FBG strain doesn't directly map to the joint angles. Currently it is black box mapping. But I plan to find a simple analytical mapping function from FBG signals to joint angles, which I hope could make MLP training simpler. 

%\mc{A kapton strip ran from each fingertip to a common wristband, as shown in Figs. \ref{fig:tacclip_design} and \ref{fig:camera_fusion} with three FBGS, located near the proximal, medial and distal joint locations. Each FBG provides a measure of the local curvature of the kapton strip.}
A polyimide strip runs from each fingertip to a common wristband, as shown in Figs.~\ref{fig:tacclip_design} and \ref{fig:camera_fusion}. Four FBGs are distributed along the strips at 3\,cm intervals. Each FBG provides a measure of the local curvature of the strip.

As illustrated in the system architecture (Fig.~\ref{fig:camera_fusion}), our approach operates through two  phases: an offline training phase for model calibration and a fusion inference phase for hand skeleton tracking.
During the offline training phase, we utilize a multi-camera motion capture system integrated with Easymocap \cite{easymocap} to acquire high-fidelity 3D hand skeletons as ground truth. By correlating the synchronized FBG strain profiles with these ground truth joint angles, we train a multi-layer perceptron (MLP) to establish a precise mapping from fiber strain to finger joint flexion.
In the fusion inference phase, a head-mounted GoPro camera is easy to deploy and serves as the primary data source for 3D hand pose estimation (e.g., via MediaPipe\cite{Lugaresi2019MediaPipeAF}%\jx{citation?}
). To handle the common challenge of visual occlusions, the framework continuously evaluates the tracking confidence score provided by the visual pipeline. When the confidence score drops below a predefined threshold, which indicates visual signal instability, the fusion module dynamically transitions from a vision-only state to a confidence-based weighted fusion. In this state, the FBG-derived flexion signals are integrated with the camera’s abduction estimations, compensating for hidden or occluded finger segments. This cross-modal fusion helps the system to maintain high-fidelity 3D hand skeleton tracking even in challenging, non-line-of-sight conditions.

Table \ref{tab:tracking_comparison} compares the results for grasping a cylinder, a small box and a thin card of known dimensions; the fingertips are modeled as cylspheres~\cite{herman1986fast} to account for their finite thickness. In these examples, the proposed vision-FBG fusion approach improves dimensional accuracy and tracking stability compared to vision-only.

%MRC this is maybe conclusions text. Doesn't belong here.
% Ultimately, this paradigm of coupling high-fidelity tactile feedback (TacClip) with vision foundation models provides a comprehensive multimodal profile of human manipulation. By unlocking the rich, fine-grained state representations that drive human decision making, this approach lays a solid foundation for robots to learn human-like manipulation strategies through imitation.  

% \mc{Can we show some result: like accuracy with vision-only (FBGs ablated) versus with everything? Explain mocap 4 cameras for ground truth. Then }

% Middle finger: 0.8cm diameter, thumb: 1.2cm diameter. 

% as noted in figure 5 

\section{Discussion and Conclusion}
\label{sec:conclusion}

We introduced TacClip, a clip-on FBG sensor that records
contact-induced fingertip deformation without placing a
sensing layer between the fingerpad and the environment.
With supervised calibration, the prototype estimates
contact force magnitude with typical errors below 0.5~N over
the tested 0--8~N range. Its 2~kHz signal also captures
dynamic events and vibrations as wearers explore textures and fine surface features.
With these capabilities, TacClip is an
adjunct rather than a complete
hand-sensing solution. It does not replace vision, pose
tracking, force/torque sensing, or distributed tactile gloves;
instead, it adds a contact channel that can reveal force changes
and dynamic tactile phenomena that those modalities may miss---and it
does so while keeping the fingerpads exposed for maximum human
acuity.  In addition, because it relies on photonic sensing, it
is unaffected by wet conditions and electromagnetic interference.

The present design does not decouple $f_x$, $f_y$,
and $f_z$ and should be interpreted as an estimator of overall
contact-load magnitude under the tested conditions. Its
response depends on finger geometry, tissue stiffness, fit,
preload, and temperature, motivating rapid per-user
calibration, improved retention, and broader testing across
users and tasks. The reported experiments establish sensing capability
rather than a complete robot-learning pipeline; the effect of
the added contact channel on downstream imitation and control
remains future work.

From an ergonomic standpoint, 
TacClip is relatively easy to don: the clip
head slides over the fingertip, requires no adhesive,
and is readily fabricated in different sizes. CAD files
and software will be provided as open-source designs
after review.
Because clip position and preload affect calibration, a snug, repeatable fit
is required. If needed, silicone pads (e.g. as used to prevent eyeglasses from sliding
down a wearer's nose) can be employed to improve fit.

At prototype scale, each wearable sensor head
costs approximately \$20 per finger, with an additional \$80 for the dorsal wristband designed for hand pose tracking, excluding the optical interrogator. The cost of an interrogator is currently high 
compared, for example, to amplifiers and analog-to-digital conversion for
strain gauges, but the cost is dropping as photonic sensing grows more popular.

\bibliographystyle{IEEEtran}
\bibliography{references/references}

\end{document}